\documentclass[11pt]{article}
\usepackage{amssymb}

\usepackage[final]{acl}

\usepackage{times}
\usepackage{latexsym}
\usepackage{amsmath}
\usepackage{booktabs}
\usepackage{amssymb}
\usepackage{adjustbox}
\usepackage{multirow}
\usepackage{array}
\usepackage[table]{xcolor}
\usepackage{pifont}
\usepackage{makecell}

\usepackage[utf8]{inputenc}
\usepackage[T1]{fontenc}
\usepackage{CJKutf8}
\usepackage[most]{tcolorbox}
\usepackage{enumitem}
\usepackage{listings}
\usepackage{fvextra}
\usepackage{multicol}
\usepackage{algorithm}
\usepackage{algpseudocode}
\usepackage{dblfloatfix}
\usepackage{tablefootnote}
\usepackage{alltt}

\algrenewcommand\algorithmicrequire{\textbf{Input:}}
\algrenewcommand\algorithmicensure{\textbf{Output:}}

\newcommand{\LineComment}[1]{%
  \Statex \hspace{\algorithmicindent}$\triangleright$~
  \begin{minipage}[t]{0.82\linewidth}
    \footnotesize #1
  \end{minipage}
}

\usepackage{fancyvrb}
\usepackage{hyperref}

\newcommand{\promptcomment}[1]{\textcolor{gray!70!black}{#1}}

\newcommand{\promptboxtitle}{}
\newcommand{\setpromptboxtitle}[1]{%
    \def\promptboxtitle{#1}%
}

\newenvironment{promptboxlist}{%
    \VerbatimEnvironment%
    \begin{tcolorbox}[
        enhanced,
        breakable,
        colback=gray!6,
        colframe=gray!65!black,
        colbacktitle=gray!65!black,
        coltitle=white,
        title={\promptboxtitle},
        fonttitle=\bfseries\small,
        boxrule=0.6pt,
        arc=2mm,
        left=6pt,
        right=6pt,
        top=4pt,
        bottom=4pt,
        before skip=8pt,
        after skip=8pt
    ]%
    \begin{Verbatim}[
        fontsize=\small,
        formatcom=\ttfamily,
        breaklines=true,
        breakanywhere=true,
        breaksymbolleft={},
        breaksymbolright={},
        tabsize=4,
        obeytabs=true,
        samepage=false,
        commandchars=\\\|\@
    ]%
}{%
    \end{Verbatim}%
    \end{tcolorbox}%
}
\usepackage{tcolorbox}
\tcbuselibrary{breakable,skins}
\usepackage{fancyvrb}

\newtcolorbox{promptbox}[1]{
    enhanced jigsaw,
    breakable,
    colback=gray!6,
    colframe=gray!65!black,
    colbacktitle=gray!65!black,
    coltitle=white,
    title={#1},
    fonttitle=\bfseries\small,
    boxrule=0.6pt,
    arc=2mm,
    left=6pt,
    right=6pt,
    top=4pt,
    bottom=4pt,
    before skip=8pt,
    after skip=8pt,
    before upper={
        \setlength{\parindent}{0pt}
        \setlength{\parskip}{0pt}
        \small
    }
}

\newcommand{\promptstep}[1]{%
    \par
    \vspace{4pt}
    \begingroup
    \setlength{\fboxsep}{2pt}%
    \noindent
    \colorbox{gray!25}{%
        \parbox{\dimexpr\linewidth-2\fboxsep\relax}{%
            \centering\bfseries\small #1%
        }%
    }%
    \endgroup
    \par
    \vspace{3pt}
}

\tcbuselibrary{breakable,listings,skins}

\newtcblisting{promptcode}[1]{
    enhanced,
    breakable,
    listing only,
    colback=gray!6,
    colframe=gray!65!black,
    colbacktitle=gray!65!black,
    coltitle=white,
    title={#1},
    fonttitle=\bfseries\small,
    boxrule=0.6pt,
    arc=2mm,
    left=6pt,
    right=6pt,
    top=4pt,
    bottom=4pt,
    listing options={
        basicstyle=\ttfamily\small,
        breaklines=true,
        columns=fullflexible
    }
}

\usepackage{fvextra}

\DefineVerbatimEnvironment{promptverbatim}{Verbatim}{
    fontsize=\footnotesize,
    breaklines=true,
    breakanywhere=true,
    breaksymbolleft={},
    breaksymbolright={},
    breakautoindent=false,
    breakindent=0pt,
    xleftmargin=0pt,
    xrightmargin=0pt
}

\usepackage[T1]{fontenc}

\usepackage[utf8]{inputenc}

\usepackage{microtype}

\usepackage{inconsolata}
\usepackage{threeparttable}

\usepackage{graphicx}

\title{DeepWeaver: Bridging the Evidence Synthesis Gap \\ in Open-Ended Question Answering}

\author{
\textbf{Xujia Wang} \quad
\textbf{Yizhe Zhang} \quad
\textbf{Bin Xu}\textsuperscript{$\dagger$} \quad
\textbf{Lei Hou} \quad
\textbf{Juanzi Li}
\\[0.6em]
Department of Computer Science and Technology, Tsinghua University
\\[0.3em]
\texttt{wang-xj22@mails.tsinghua.edu.cn}
}

\begin{document}
\maketitle
\begin{abstract}
Retrieve-then-generate pipelines are commonly used to produce deep-research answers for open-ended questions, but retrieval alone is insufficient: LLMs must organize noisy and fragmented evidence into comprehensive, well-cited answers. We refer to this process as \textbf{evidence synthesis}. However, direct generation often underuses evidence, misaligns citations, and collapses diverse information into shallow summaries, exposing an evidence synthesis gap between retrieval and generation. Thus, we propose \textbf{DeepWeaver}, a novel framework that weaves noisy retrieved evidence into comprehensive answers by maintaining Thought Block Chains (TBCs), a structured representation that groups claims, salient information, keywords, and supporting evidence. DeepWeaver uses subordinate TBCs to inspect residual evidence, commit TBC revisions, and discover new claims before final generation. We evaluate DeepWeaver on open-ended QA over both knowledge bases and the web, and introduce \textbf{LoQA}, a high-density benchmark for evidence synthesis. Across LLMs, DeepWeaver improves content sufficiency, citation grounding, and detail preservation on LoQA, while achieving deeper insights and higher citation quality on DeepResearch Bench. These results show that evidence weaving is effective for bridging retrieval and generation in open-ended QA. Our code is available at \href{https://github.com/KlozeWang/DeepWeaver}{this URL}.
\end{abstract}

\section{Introduction}
\begingroup
\renewcommand{\thefootnote}{\fnsymbol{footnote}}
\footnotetext[2]{\quad  Corresponding author.}
\endgroup
Large language models (LLMs) are increasingly being used to generate research-style answers grounded in external evidence \cite{menick2022teaching, asai2024openscholar}. For domain-specific open-ended questions and deep-research-style tasks, answering a query may require integrating evidence from hundreds of retrieved passages, web pages, books, or technical documents.  As a result, evidence retrieval and synthesis are central to both retrieval-augmented generation (RAG) pipelines \citep{lewis2021retrievalaugmentedgenerationknowledgeintensivenlp, izacard2021leveragingpassageretrievalgenerative} and web-based deep-research agents \citep{nakano2021webgpt, yao2022react}. These systems share a common workflow: retrieve evidence, place it into the context, and prompt the model to synthesize a comprehensive answer \cite{huang2024survey}.

\begin{figure}[t]   
  \centering           
  \includegraphics[width=\linewidth]{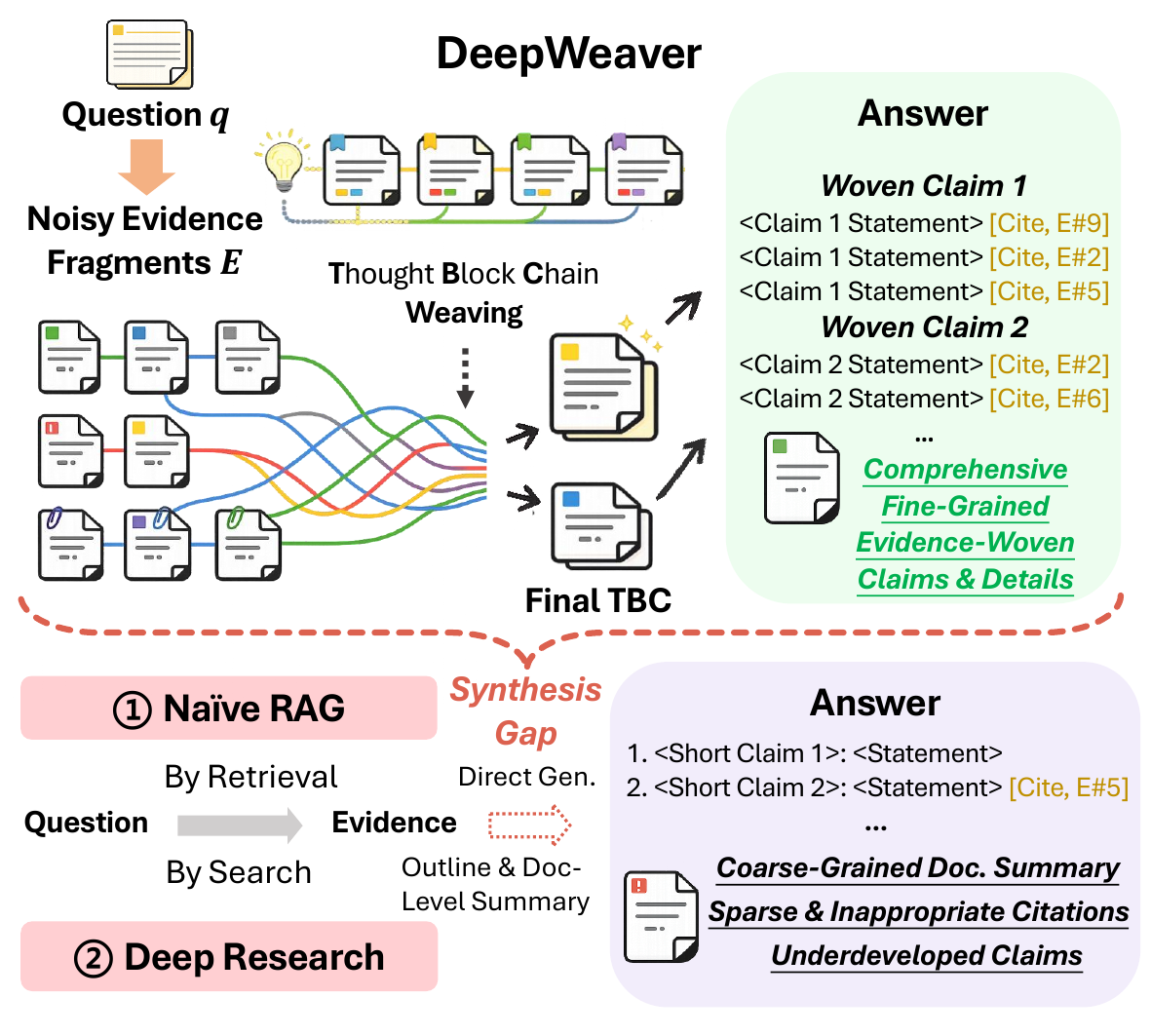} 
  \vspace{-17.5pt}
  \caption{DeepWeaver bridges the evidence synthesis gap between retrieval and generation by weaving noisy evidence into structured Thought Block Chains (TBCs). Unlike naïve retrieval-augmented generation, DeepWeaver synthesizes noisy evidence fragments into answers with well-supported woven claims, fine-grained details, and grounded citations.}
  \label{fig:intro}
  \vspace{-17.5pt}
\end{figure}

Although prior work has extensively studied how to improve evidence retrieval quality \citep{han2024retrieval, guo2024lightrag, sun2025dynamicragleveragingoutputslarge}, retrieval itself is merely the first step: \textbf{enabling LLMs to effectively use \textit{extensive, noisy, and fragmented} evidence to generate comprehensive answers remains a key challenge}. To better integrate upstream evidence, context-refinement methods \citep{xu2023recomp, jin2025hierarchical} compress, filter, or restructure passages before generation, while divide-and-conquer methods \citep{zhang2024chain,zhou2024llm} split long contexts into segments and aggregate intermediate outputs. End-to-end deep-research systems \citep{wang2024autosurvey,tao2025webshaper,li2025webweaver} go further by automating evidence organization and long-form answer writing with agent memory and outline-guided generation.



However, prior methods overestimate LLMs' ability to fully exploit input evidence to derive in-depth, diverse claims. By primarily optimizing evidence presentation and outline-guided writing \citep{xu2025comprehensivesurveydeepresearch}, these approaches often fall short in deeply synthesizing retrieved evidence into well-developed answers. When faced with noisy, fragmented, and knowledge-dense evidence, relying solely on the LLM's internal reasoning often obscures fine-grained details and produces underdeveloped claims. This disconnect between raw retrieved evidence and comprehensive generation is what we define as \textbf{the evidence synthesis gap}.

We argue that \emph{an intermediate module for orchestrating evidence and weaving evidence-grounded claims} is essential for bridging the gap between retrieval and final answer generation (Figure \ref{fig:intro}). The central idea is to structure noisy retrieved evidence into fine-grained thought units, rather than synthesizing answers in a single pass.

To this end, we propose \textbf{DeepWeaver}, a novel framework for open-ended answer generation grounded in noisy evidence. DeepWeaver maintains a \textbf{T}hought \textbf{B}lock \textbf{C}hain (TBC) structure, where each thought block corresponds to a discovered aspect of the answer and stores its claim, keywords, salient information, and supporting evidence fragments. DeepWeaver weaves TBCs by identifying overlooked evidence in the current TBC and generating subordinate TBCs to inspect residual evidence that has not been sufficiently covered. Newly discovered information and claims are then committed back into the main TBC. This design produces evidence-woven claims and links these claims to relevant fragments, thereby improving answer comprehensiveness and citation quality while alleviating context-window pressure.

We evaluate DeepWeaver on our \textbf{LoQA} benchmark (\S \ref{sec:loqa}), an open-ended answer generation benchmark with extensive noisy evidence. LoQA contains 100 water-environment research questions, each paired with evidence retrieved from a Chinese expert knowledge base comprising 500 books and over 100M characters. It stress-tests whether QA systems can comprehensively use evidence, preserve domain-specific details, and generate well-cited answers. We also evaluate DeepWeaver on web-based deep-research questions using \textbf{DeepResearch Bench} \cite{du2025deepresearch}, which contains PhD-level research tasks across diverse fields. Experiments show that DeepWeaver substantially improves content sufficiency, citation grounding, and detail preservation on LoQA, while achieving deeper insights and higher citation quality than advanced agents on DeepResearch Bench.

Our main contributions include:

\begin{itemize}
    \item We identify the evidence synthesis gap in open-ended QA, highlighting the need for an explicit synthesis stage beyond retrieval.
    
    \item We propose DeepWeaver, a framework that orchestrates noisy retrieved evidence into structured Thought Block Chains and generates open-ended answers via evidence weaving.
    
    \item We introduce LoQA, a benchmark for evaluating evidence-grounded answer generation. Experiments show that DeepWeaver improves answer comprehensiveness and citation quality on both LoQA and DeepResearch Bench.
\end{itemize}

\section{Task Formulation}




We evaluate whether a QA system can synthesize a large set of evidence fragments into comprehensive open-ended answers. This setting reflects realistic information-seeking needs: answering a query may involve multiple claims, while the supporting evidence is scattered across sources (\textit{e.g.}, books, technical documents, and passages). Although RAG techniques can retrieve such evidence, whether QA systems can fully exploit it is unclear.

\subsection{Definition}
\label{sec:def}

Given an open-ended research-style question $q$ and a retrieved evidence pool
$E = \{e_1, e_2, \ldots, e_N\}$ collected from relevant documents, a QA system is required to generate an answer $y$ that thoroughly resolves $q$ while being grounded in the evidence. Answers are evaluated according to three aspects that characterize evidence synthesis quality:

\begin{itemize}
    \item \textbf{Content sufficiency}: the answer should cover the major content and claims implied by the question and synthesized from the evidence.
    \vspace{-4pt}
    \item \textbf{Detail preservation}: the answer should retain fine-grained information rather than collapse it into generic summaries.
    \vspace{-4pt}
    \item \textbf{Citation grounding}: the answer should ground its claims in relevant evidence.
\end{itemize}

\subsection{The LoQA Benchmark}
\label{sec:loqa}

LoQA is built from a Chinese water-environment domain knowledge base containing 500 books and over 100M characters. We choose this domain because it contains dense technical knowledge, complex causal relations, and practical open-ended questions, which involve evidence across multiple subtopics, such as pollution control, water quality assessment, and watershed management.

\paragraph{Construction.} LoQA contains 100 open-ended research-style questions. We generate 2,700 candidate questions using multiple state-of-the-art LLMs based on book contexts, and then filter them using LLM-based scores for retrieval comprehensiveness, concreteness, domain depth, and expression completeness. We retrieve evidence for each candidate and select the final questions with the highest average similarity to the retrieved fragments, ensuring that each question involves rich evidence (\S \ref{sec:loqa_con}).

About 100 evidence fragments (chunked into 1,024-token segments) are retrieved for each question. To simulate a noisier context, we add another 100 random evidence fragments\footnote{Each question is paired with about $200\times 1024= 205\text{K}$ tokens of evidence, approaching the context window of many LLMs.}.

Gold answers for long-context, open-ended, domain-specific questions are difficult to produce and evaluate. Inspired by prior work \cite{shao2024assisting}, we therefore evaluate evidence synthesis quality by assessing answer comprehensiveness along several controlled dimensions. Specifically, LoQA decomposes this subjective notion into three dimensions: (1) \textbf{citation grounding}, (2) \textbf{content sufficiency}, and (3) \textbf{detail preservation}.

\paragraph{Citation grounding.}
We use a strong LLM judge, DeepSeek-V3.2, to determine whether each evidence fragment is relevant to the question. 
This splits the evidence pool $E$ into a relevant set $E_R$ and an irrelevant set $E_I$. 
Given a generated answer $y$, let $C$ denote the set of evidence fragments cited by $y$. 
We compute three citation metrics: (1) \textbf{Citation Count (CC)}: $|C|$, (2) \textbf{Relevant Count (RC)}: $|C \cap E_R|$, and (3) \textbf{Relevant Ratio (RR)}: $|C \cap E_R|/|C|$.
These metrics measure whether the system actively uses evidence and properly avoids citing or integrating irrelevant evidence.

\paragraph{Content sufficiency.} This dimension evaluates whether the answer sufficiently covers the key terms and arguments synthesized from the evidence. To obtain reference answers, we use DeepSeek-V3.2 to generate five answers from different partitions of the relevant evidence set $E_R$:
\[
E_R^{0\text{-}25\%}, \ 
E_R^{25\text{-}50\%}, \ 
E_R^{50\text{-}75\%}, \ 
E_R^{75\text{-}100\%}, \ 
E_R^{0\text{-}100\%}.
\]
This yields five reference answers $y_1, y_2, y_3, y_4, y_5$.

We compute two lexical-level metrics:

(1) \textbf{Recall}, the average of the longest common subsequence (LCS) ratio:
\vspace{-6pt}
    \[
    \frac{1}{5}\sum_{i=1}^{5} \frac{\mathrm{LCS}(y_i, y)}{|y_i|}.
    \]
(2) \textbf{Word Alignment (WA)}, the average BLEU-4 score, adopting the generated answer $y$ as the target and each reference answer $y_i$ as the source:
\vspace{-6pt}
    \[
    \frac{1}{5}\sum_{i=1}^{5} \mathrm{BLEU\text{-}4}(y_i, y).
    \]
We also compute the \textbf{Argument Sufficiency (AS)} metric, using an LLM to extract a set of atomic arguments $A$ from reference answers $y_1,\ldots,y_5$. 
Then we judge whether each argument in $A$ is covered by the generated answer $y$ via LLM:
$$
\frac{1}{|A|}\sum_{a \in A} \mathbb{I}(a \text{ is covered by } y).
$$

\paragraph{Detail preservation.}
We evaluate whether the generated answer preserves fine-grained details from the evidence pool. 
We first use an LLM to create cloze-style questions from the answer $y_5$. 
The blanks mainly target technical terms, concepts, and short phrases. We judge whether each blank can be correctly recovered by providing the generated answer $y$ as the reference document. 
\textbf{Detail Preservation (DP)} is the proportion of cloze blanks in $B$ that can be correctly recovered:
\[
\frac{1}{|B|}\sum_{b \in B} \mathbb{I}(b \text{ can be recovered from } y).
\]
\begin{figure*}[t]
    \centering
    \includegraphics[width=0.985\linewidth]{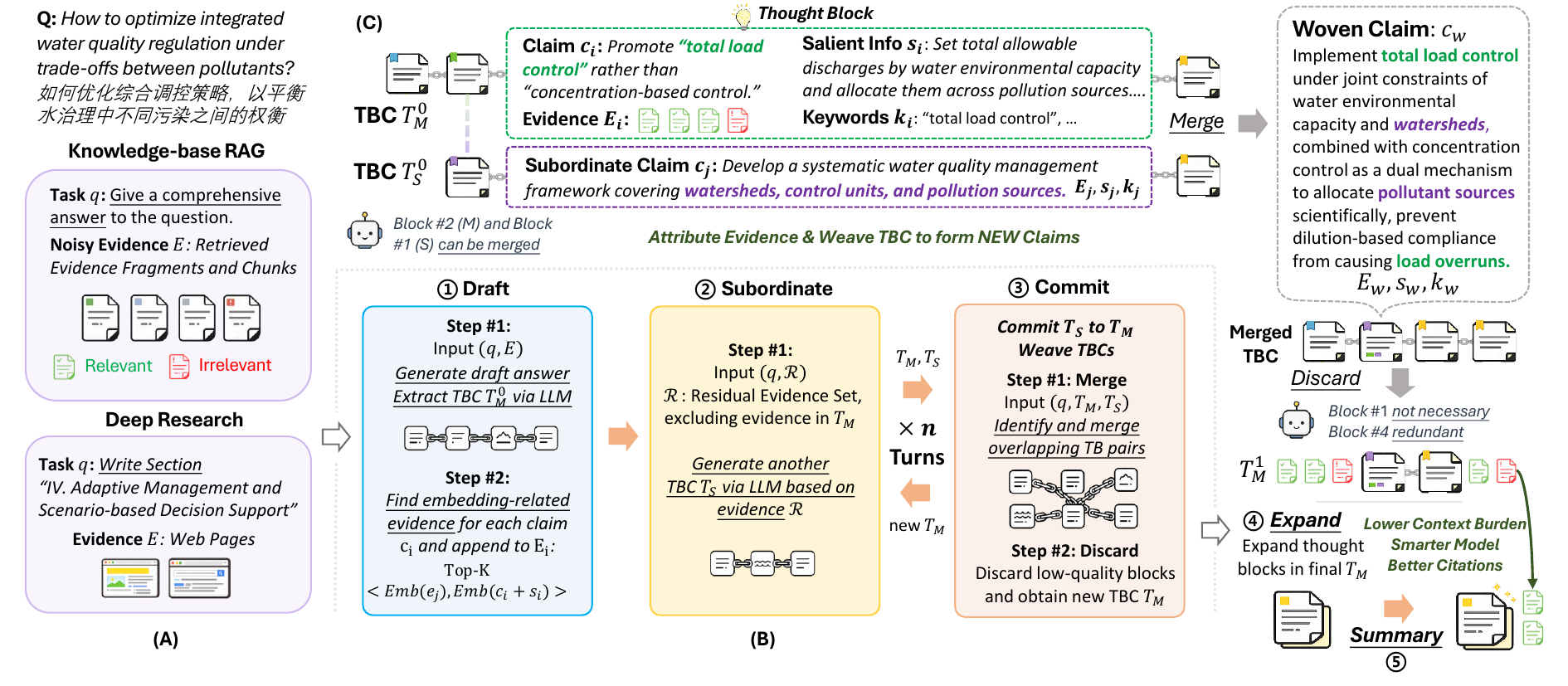} 
    \vspace{-10pt}
    \caption{\textbf{Overview of DeepWeaver.} 
(A) We apply DeepWeaver to two typical retrieve-then-generate scenarios. 
(B) DeepWeaver maintains and refines the TBC through three stages: Draft, Subordinate, and Commit. 
(C) A concrete example of commitment shows how DeepWeaver synthesizes new claims from residual evidence by TBC weaving.}
    \vspace{-12.5pt}
    \label{fig:method}
\end{figure*}

\section{Method}

\subsection{Thought Block Chain}

We define the \textbf{T}hought \textbf{B}lock \textbf{C}hain (TBC) as an explicit data structure that bridges retrieved evidence and final answer generation. The TBC serves as the core information structure maintained by DeepWeaver, decomposing the answer into a sequence of thought blocks:
\[
\mathcal{T} = [b_1, b_2, \ldots, b_K].
\]
Each block $b_i$ encapsulates a candidate claim and stores the corresponding keywords, salient information, and supporting evidence fragments:
\[
b_i = (c_i, k_i, s_i, E_i),
\]
where $c_i,k_i$ and $s_i$ denote the claim, keywords, and salient information, respectively, and $E_i \subseteq E$ is the subset of evidence fragments linked to the claim.


Unlike a plain outline that tends to provide a survey rather than a question-resolving answer, the TBC organizes detailed evidence into fine-grained woven claims, while maintaining an explicit mapping between claims and relevant evidence and reducing the context burden during generation.

\subsection{Evidence Weaving}
\label{sec:weaving}

The core mechanism of DeepWeaver (Figure \ref{fig:method}(B)) lies in the maintenance of a main TBC $\mathcal{T}_M$. It updates $\mathcal{T}_M$ through multi-stage refinement. Overall, DeepWeaver consists of three ordered weaving stages: Draft, Subordinate, and Commit.


\paragraph{Draft.}
Given the question $q$ and the retrieved evidence pool $E$, the LLM first produces an answer draft via direct generation. DeepWeaver then extracts the \textbf{initial main TBC $\mathcal{T}_M^0$} from this draft. This step leverages the LLM's global view of the evidence pool, allowing it to form an initial set of claims from a broad perspective.



\paragraph{Subordinate.}
The initial TBC $\mathcal{T}_M^0$ may be incomplete, as direct generation can overlook useful evidence and omit important claims. The subordinate stage is designed to identify evidence fragments that are neglected, weakly covered, or insufficiently reflected in the current TBC. We define the \textbf{residual evidence set} $\mathcal{R}_0$ of the initial TBC $\mathcal{T}_M^0$ as:
\[
\mathcal{R}_0 = \{e_j \in E \mid e_j \text{ is not covered by } \mathcal{T}_M^0\}.
\]

We say an evidence fragment is covered by the initial TBC $\mathcal{T}_M^0$ if it satisfies either: (1) it is directly mentioned in $\mathcal{T}_M^0$; or (2) it is among the top-$k$ fragments most similar to the string $c_i+s_i$ in the embedding space, and the similarity is higher than the average similarity of the evidence $E_i$ in the corresponding block $b_i\in \mathcal{T}_M^0$. The complement of the covered evidence set is defined as $\mathcal{R}_0$.

DeepWeaver generates a \textbf{subordinate TBC $\mathcal{T}_S^0$} (Figure \ref{fig:method} \ding{173}) from the evidence in $\mathcal{R}_0$. This procedure acts as a local inspector of overlooked evidence. It discovers missing aspects, additional details, and alternative supporting evidence that are not captured by the main TBC.

\paragraph{Commit.}
The useful information in $\mathcal{T}_S^0$ is then committed back to the main TBC $\mathcal{T}_M^0$ through two operations: \textbf{Merge} and \textbf{Discard} (Figure \ref{fig:method} \ding{174}). 

The merge operation identifies overlapping claim pairs between the two TBCs and merges their claims $c_i$ and keywords $k_i$, while their salient information $s_i$ and evidence sets $E_i$ are combined to form a merged block. \textit{This produces woven claims that integrate multi-dimensional explanations.}

After merging, the discard operation removes irrelevant, redundant, or weakly supported claims with respect to the question, preventing TBC from over-expanding. This yields a \textbf{refined TBC $\mathcal{T}_M^1$}:
\[
\textsc{Discard}\big(q, \textsc{Merge}(\mathcal{T}_M^0,\mathcal{T}_S^0)\big)
\rightarrow \mathcal{T}_M^1.
\]

DeepWeaver repeats the subordinate and commit stages for $n$ rounds to refine the woven claims and incorporate additional supporting evidence\footnote{Note that TBC maintenance is lightweight, adding minor output-token overhead compared with answer generation.}:
\[
\mathcal{T}_M^0 \rightarrow \mathcal{T}_M^1 \rightarrow \cdots  \rightarrow  \mathcal{T}_M^t  \rightarrow  \cdots \rightarrow \mathcal{T}_M^n.
\]

Furthermore, to reduce the context burden during TBC construction, the draft and subordinate stages randomly sample $r\ (r < |E|)$  evidence fragments from $E$ and $\mathcal{R}_t$ (the residual evidence set at refinement turn $t$), respectively. Across multiple revision rounds, the model can inspect the full evidence pool while avoiding the burden of processing an excessively long context in one pass.

\subsection{Evidence-Grounded Answer Generation}

After evidence weaving, DeepWeaver generates the final answer from the refined TBC $\mathcal{T}_M^n$ (Figure~\ref{fig:method} \ding{175}--\ding{176}). For each block $b_i$, the model receives its metadata and linked evidence subset $E_i$ as a focused local context, and then generates an \textbf{answer section $S_i$} grounded in the corresponding evidence:
\[
S_i = \textsc{Generate}(q, c_i, s_i, E_i).
\]
The final answer\footnote{In this work, we focus on evidence synthesis quality and answer comprehensiveness. In settings such as DeepResearch Bench, where basic readability is also required, we add simple optional steps including (1) grouping TBC blocks into article sections, (2) polishing the generated sections, and (3) adding introduction and conclusion paragraphs.} is composed of all generated sections using the LLM to build links among $S_i$:
\[
y_i = \textsc{Append} (y_{i-1},S_i),\ y=y_K
\]

This block-wise generation strategy has two advantages: (1) it decomposes context pressure into claim-level generation, and (2) it improves citation grounding because each section is generated from a smaller and more relevant evidence subset.

\section{Experiments}

We evaluate DeepWeaver on both knowledge-base open-ended question answering and end-to-end web-based open-ended deep research.

\subsection{Datasets}

\paragraph{LoQA.}
A high-density evidence-based QA benchmark (\S \ref{sec:loqa}) with 100 research-style practical water-environment questions. Each question is paired with long-context evidence retrieved from a 500-book knowledge base, evaluating whether systems can deeply synthesize fragmented retrieved evidence to generate comprehensive answers.

\paragraph{DeepResearch Bench.}
It contains 100 PhD-level deep-research tasks across 22 fields, where evaluated agents search for web information and write long-form answers. The \textbf{RACE} split measures answer quality using LLM-judged criteria, including comprehensiveness, insight, instruction following, and readability\footnote{Here, ``comprehensiveness'' and ``insight'' also correspond to two aspects of  evidence synthesis quality under our definition (\S~\ref{sec:def}).}, while \textbf{FACT} measures citation accuracy and the number of effective citations.

\subsection{Baselines}

We compare DeepWeaver with representative evidence-based QA baselines on LoQA.

\paragraph{RAG.}
An LLM directly generates the answer from evidence. We consider three context settings: the full evidence pool $E$, the relevant evidence set $E_R$, and a randomly sampled subset $E_{120}$ containing 120 chunks \citep{lewis2021retrievalaugmentedgenerationknowledgeintensivenlp}.
\vspace{-2.5pt}

\paragraph{Skeleton-of-Thought (SoT).}
The model first generates a thought outline and then writes the answer section by section, with the full evidence pool provided in the input context \citep{ning2024skeleton}.
\vspace{-2.5pt}

\paragraph{Plan-and-Solve (PaS).}
The model generates a plan to produce the answer \citep{wang2023plan}.
\vspace{-2.5pt}

\paragraph{Chain-of-Agents (CoA).}
A divide-and-conquer framework that splits the evidence pool $E$ into smaller subsets, processes them batch by batch, and gradually refines the answer \citep{zhang2024chain}.
\vspace{-2.5pt}

\paragraph{LongRefiner.}
A method that preprocesses the evidence pool $E$ by extracting structured articles and sentences from noisy evidence \citep{jin2025hierarchical}.

For DeepResearch Bench, baselines include advanced agents\footnote{doubao-research, kimi-research, Claude-research, and openai-deepresearch.} and WebWeaver \citep{li2025webweaver}, a dual-agent framework that uses a planner to search the web and write outlines, and a writer to fulfill the outline using a memory bank.
\newcommand{\modelrow}[1]{%
\noalign{\vskip 3pt}
\multicolumn{8}{l}{\itshape\hspace{-10pt} $\blacktriangleright$ \textbf{#1}} \\
\noalign{\vskip 3pt}

}
\newcommand{\modelrowt}[1]{%
\multicolumn{7}{c}{\itshape\hspace{-10pt} $\blacktriangleright$ \textbf{#1}}
}

\newcommand{\modelrowtm}[1]{%
\multicolumn{8}{c}{\itshape\hspace{-10pt} $\blacktriangleright$ #1} \\
\noalign{\vskip 2pt}
}

\newcommand{\green}[1]{\textcolor{green!50!black}{#1}}

\begin{table*}[htbp]
\centering
\setlength{\tabcolsep}{4.8pt}
{\fontsize{9pt}{10pt}\selectfont
\renewcommand{\arraystretch}{0.95}
\begin{adjustbox}{max width=\textwidth}
\begin{tabular}{lccc!{\hspace{6pt}}ccc!{\hspace{2pt}}c|ccc!{\hspace{6pt}}ccc!{\hspace{2pt}}c}
\toprule
\multirow{3}{*}{\textbf{Method}}
& \modelrowt{Qwen3-30B-A3B-Instruct-2507}
& \modelrowt{Qwen3-30B-A3B-Thinking-2507} \\
\cmidrule(lr){2-8} \cmidrule(lr){9-15}
& \multicolumn{3}{c}{\textbf{Content S.}}
& \multicolumn{3}{c}{\textbf{Citation G.}}
& \textbf{Detail}
& \multicolumn{3}{c}{\textbf{Content S.}}
& \multicolumn{3}{c}{\textbf{Citation G.}}
& \textbf{Detail} \\
\cmidrule(lr){2-4} \cmidrule(lr){5-7} \cmidrule(lr){8-8}
\cmidrule(lr){9-11} \cmidrule(lr){12-14} \cmidrule(lr){15-15}
& \textbf{Recall} & \textbf{WA} & \textbf{AS}
& \textbf{CC} & \textbf{RC} & \textbf{RR} & \textbf{DP}
& \textbf{Recall} & \textbf{WA} & \textbf{AS}
& \textbf{CC} & \textbf{RC} & \textbf{RR} & \textbf{DP} \\
\midrule
$E_{120}$-RAG
& 20.4 & 4.0 & 70.6 & 17.7 & 13.6 & 78.9 & 56.4
& 24.0 & 5.6 & 64.8 & 12.8 & 10.8 & 85.0 & 51.0 \\

$E$-RAG
& 19.8 & 4.0 & 68.0 & 11.9 & 8.8 & 74.3 & 52.9
& 21.9 & 5.0 & 59.9 & 10.6 & 9.0 & 75.3 & 53.9 \\

$E_R$-RAG
& 21.8 & 4.5 & 76.5 & 21.4 & 21.4 & \underline{100} & 58.8
& 24.6 & 5.9 & 68.0 & 16.3 & 16.3 & \underline{100} & 55.9 \\

\midrule
SoT
& 20.3 & 4.2 & 73.4 & 13.8 & 7.7 & 53.6 & 58.7
& 30.1 & \textbf{8.2} & 61.8 & 24.2 & 15.4 & 64.5 & 29.5 \\

PaS
& 13.8 & 2.0 & 58.4 & 9.9 & 4.8 & 48.4 & 44.8
& 10.9 & 2.0 & 33.2 & 6.5 & 4.0 & 35.8 & 46.3 \\

Chain-of-Agents
& 14.7 & 2.4 & 70.3 & 19.2 & 15.3 & 81.2 & 60.2
& 8.2 & 1.1 & 34.4 & 8.6 & 7.8 & 91.0 & 30.3 \\

LongRefiner
& 21.0 & 4.0 & 78.2 & 21.1 & 16.7 & 79.7 & 57.6
& 23.9 & 5.4 & 69.0 & 16.7 & 14.7 & 87.5 & 36.4 \\

\midrule
\textbf{DeepWeaver}
& \textbf{27.4} & \textbf{6.1} & \textbf{83.5} & \textbf{26.6} & \textbf{23.5} & \textbf{88.8} & \textbf{69.5}
& \textbf{34.5} & 8.0 & \textbf{85.1} & \textbf{30.9} & \textbf{28.6} & \textbf{92.3} & \textbf{69.1} \\
\green{$\Delta$ over E-RAG}
& \green{+7.6} & \green{+2.1} & \green{+15.5} & \green{+14.7} & \green{+14.7} & \green{+14.5} & \green{+16.6}
& \green{+12.6} & \green{+3.0} & \green{+25.2} & \green{+20.3} & \green{+19.6} & \green{+17.0} & \green{+15.2} \\
\bottomrule
\end{tabular}
\end{adjustbox}
}
\vspace{-5pt}
\caption{Comprehensiveness of answers generated by evidence-based QA systems on LoQA, reflecting  evidence synthesis quality. Results are shown in terms of content sufficiency (Content S.), citation grounding (Citation G.) and detail preservation (Detail). Metric definitions are provided in \S\ref{sec:loqa}. The best results are highlighted in \textbf{bold}.}
\label{tab:loqa_main}
\vspace{-12.5pt}
\end{table*}

\subsection{Implementation Details}

We evaluate DeepWeaver under multiple backbones, including Qwen3-30B-A3B-Instruct-2507, Qwen3-30B-A3B-Thinking-2507 \citep{yang2025qwen3}, DeepSeek-V3.2, DeepSeek-V4-Flash \cite{liu2024deepseek,deepseekai2026deepseekv4}, and Qwen3.5-122B-A10B \citep{qwen3.5}. We set refinement rounds to $n=2$, the evidence retrieval size to $k=5$, and the number of randomly sampled chunks to $r=120$. We use GLM-Embedding-3 \citep{5team2025glm45agenticreasoningcoding} to compute text embeddings. The maximum generation length is 32,768 tokens.

For LoQA evaluation, we use DeepSeek-V3.2 as the judge for Argument Sufficiency (AS) and Detail Preservation (DP), with temperature set to 0. For DeepResearch Bench, we follow the official setting and use Gemini-2.5-Pro \cite{comanici2025gemini} for RACE and Gemini-2.5-Flash for FACT evaluation. More implementation details for DeepResearch Bench are provided in Appendix~\S\ref{sec:drsetting}.

\subsection{Main Results}
Table~\ref{tab:loqa_main} reports the main results on LoQA. DeepWeaver consistently improves evidence
synthesis quality across both instruct and thinking backbone models. \textbf{DeepWeaver substantially outperforms direct generation and evidence-based QA baselines.} With Qwen3-30B-A3B-Instruct-2507, DeepWeaver surpasses $E$-RAG by 15.5\% in Argument Sufficiency, +14.7 Relevant Citations, 14.5\% in Relevant Citation Ratio, and 16.6\% in Detail Preservation. These gains show that DeepWeaver improves answer quality across multiple dimensions, including content sufficiency, citation grounding, and detail preservation.

\textbf{We further highlight the evidence synthesis gap: direct generation is insufficient for using noisy RAG evidence.} For both backbones, $E$-RAG does not outperform $E_{120}$-RAG, despite receiving more evidence. This suggests that adding more retrieved chunks can increase the context burden and cause the model to overlook useful information. The oracle $E_R$-RAG setting receives only relevant evidence, yet it still lags behind DeepWeaver. This indicates that removing noise alone is insufficient to solve the evidence synthesis problem.

Among the reasoning and divide-and-conquer baselines, SoT struggles to preserve fine-grained details and identify relevant evidence under heavy context burden. PaS is less effective because open-source models often struggle to produce reliable writing plans for complex, evidence-heavy questions. CoA and LongRefiner improve citation usage in some cases by reducing context burden or refining evidence, but they still suffer from insufficient claims and missing details. These results support our main claim: \textbf{comprehensive answer generation requires not only retrieving or filtering evidence, but also an effective evidence synthesis process} that weaves fine-grained evidence into coherent claim-level structures.

\subsection{Ablation Study}


We ablate two core mechanisms of DeepWeaver on Qwen3-30B-A3B-Instruct-2507: TBC subordination and commitment (\S \ref{sec:weaving}). In \textit{w/o subordinate}, the system directly expands the initial TBC $\mathcal{T}_M^0$ into the final answer. In \textit{w/o commit}, the system appends the subordinate TBC $\mathcal{T}_S^0$ to $\mathcal{T}_M^0$, without the merge and discard operations.

\newcommand{\redtext}[1]{\textcolor{red}{#1}}
\begin{table}[htbp]
\vspace{-7.5pt}
\centering
\setlength{\tabcolsep}{3pt}
{\fontsize{10pt}{10pt}\selectfont
\renewcommand{\arraystretch}{1.1}
\begin{adjustbox}{max width=\columnwidth}
\begin{tabular}{lcccccc}
\toprule
& \textbf{AS}
& \textbf{RC} & \textbf{RR} &  \textbf{DP} & \textbf{\#Blocks}\\
\midrule
\textbf{DeepWeaver} &  83.5 & \textbf{23.5} & 88.8 & \textbf{69.5} & 6.63\\
\quad w/o TBC subordinate & \redtext{78.6} & 21.2 & \textbf{89.2} &  \redtext{62.4} & 5.76 \\
\quad w/o TBC commit & \textbf{84.4} & 22.3 & \redtext{85.8} & 67.9 & \redtext{10.2} \\
\bottomrule
\end{tabular}
\end{adjustbox}
}
\vspace{-7.5pt}
\caption{Ablation study on TBC subordination and commitment. Scores in \redtext{red} highlight variants with severe performance degradation.}
\vspace{-12.5pt}
\label{tab:abla}
\end{table}

Table~\ref{tab:abla} shows that removing subordination hurts all three dimensions of evidence synthesis quality, indicating that the initial TBC is often incomplete and requires further refinement to recover overlooked evidence. Removing commitment makes answers less concise and weakens claim weaving: it achieves comparable AS and DP while producing more blocks, suggesting that it introduces overlapping and redundant claims.

\begin{figure}[t]
    \centering
    \includegraphics[width=0.95\linewidth]{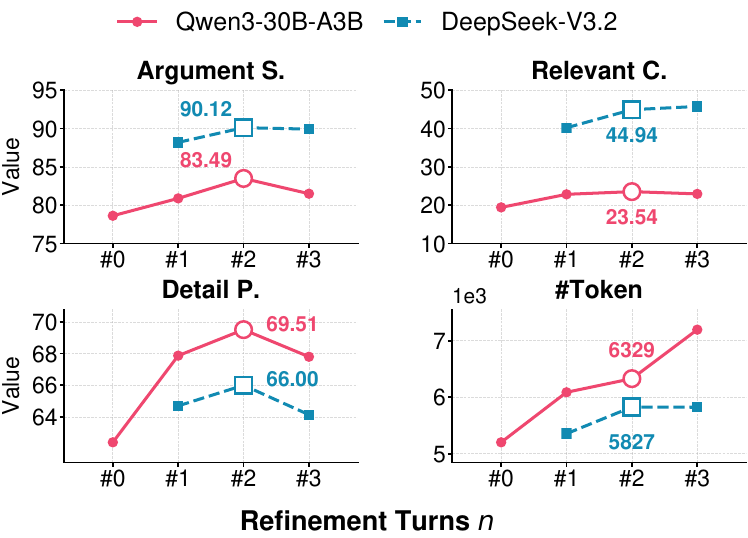} 
    \vspace{-5pt}
    \caption{Effect of refinement rounds on final performance. Increasing the number of refinement rounds from $n=1$ to $n=2$ improves answer comprehensiveness, while a third round brings no clear additional gain. $n=2$ provides the best cost--performance trade-off.}
    \label{fig:turns}
    \vspace{-17.5pt}
\end{figure}

We also study the optimal refinement rounds $n$. As shown in Figure~\ref{fig:turns}, DeepWeaver remains stable across different values of $n$, with $n=2$ achieving the best overall performance. Increasing the number of rounds from one to two improves content sufficiency, citation grounding, and detail preservation. This indicates that the second subordinate-and-commit step helps enrich the claims and final TBC. A third round yields no substantial improvements, possibly because excessive refinement introduces marginally useful information. Considering both performance and efficiency, we set $n=2$ by default. \textbf{With only two rounds of refinement over the initial TBC, DeepWeaver brings large gains.} This demonstrates the efficiency of the evidence weaving mechanism.


\subsection{Cross-Model Generalizability}

Table~\ref{tab:multim} evaluates whether DeepWeaver generalizes across different LLM backbones. The results show that DeepWeaver brings consistent gains across backbone models. For example, on DeepSeek-V4-Flash, DeepWeaver improves Recall from 24.2 to 31.6, AS from 75.4\% to 86.5\%, RC from 31.0 to 44.6, and DP from 60.0\% to 67.4\%.

These results indicate that the \textbf{benefit of DeepWeaver does not depend on a specific backbone model}. Instead, evidence weaving provides a general mechanism for improving the use of noisy evidence. Notably, DeepWeaver with Qwen3-30B-A3B-Instruct-2507 outperforms direct generation with DeepSeek-V3.2, even though DeepSeek-V3.2 is a strong long-context model. This shows that explicitly weaving evidence is more effective than relying solely on stronger long-context capabilities.

\begin{table}[htbp]
\centering
\setlength{\tabcolsep}{4.8pt}
{\fontsize{9pt}{10pt}\selectfont
\renewcommand{\arraystretch}{0.95}
\begin{adjustbox}{max width=\columnwidth}
\begin{tabular}{lccc!{\hspace{6pt}}ccc!{\hspace{2pt}}c}
\toprule
\multirow{3}{*}{\textbf{Method}}
& \multicolumn{3}{c}{\textbf{Content S.}}
& \multicolumn{3}{c}{\textbf{Citation G.}}
& \textbf{Detail}\\
\cmidrule(lr){2-4} \cmidrule(lr){5-7} \cmidrule(lr){8-8} 
& \textbf{Recall} & \textbf{WA} & \textbf{AS}
& \textbf{CC} & \textbf{RC} & \textbf{RR} & \textbf{DP} \\
\midrule 
\modelrowtm{DeepSeek-V4-Flash}
w/o DW & 24.2 & 6.1 & 75.4 & 35.4 & 31.0 & 88.2 & 60.0 \\
\textbf{w DW}
& \textbf{31.6} & \textbf{10.6} & \textbf{86.5} & \textbf{50.5} & \textbf{44.6} & \textbf{88.4} & \textbf{67.4} \\ \midrule
\modelrowtm{Qwen3.5-122B-A10B}
w/o DW & 21.7 & 4.6 & 77.7 & 32.4 & 28.9 & 89.2 & 53.5 \\
\textbf{w DW}
& \textbf{29.0} & \textbf{7.3} & \textbf{87.7} & \textbf{49.9} & \textbf{44.4} & \textbf{89.3} & \textbf{64.6} \\ \midrule
\modelrowtm{DeepSeek-V3.2}
w/o DW
& 24.5 & 5.8 & 78.5 & 27.0 & 21.7 & 81.7 & 59.6 \\
\textbf{w DW}
& \textbf{32.4} & \textbf{10.3} & \textbf{90.1} & \textbf{51.8} & \textbf{44.9} & \textbf{86.7} & \textbf{65.7} \\ \midrule
\modelrowtm{Qwen3-30B-A3B-Instruct-2507}
\green{w DW} & \green{27.4} & \green{6.1} & \green{83.5} & \green{26.6} & \green{23.5} & \green{88.8} & \green{69.5} \\
\bottomrule
\end{tabular}
\end{adjustbox}
}
\vspace{-5pt}
\caption{Generalizability of DeepWeaver across recent advanced LLMs. DeepWeaver consistently yields substantial gains across different backbone models. Notably, DeepWeaver with Qwen3-30B-A3B-Instruct-2507 substantially outperforms direct generation with DeepSeek-V3.2, a strong long-context model.}
\label{tab:multim}
\vspace{-15pt}
\end{table}

\begin{table*}[t]
\centering
\setlength{\tabcolsep}{8.0pt}
{\fontsize{8.5pt}{11pt}\selectfont
\renewcommand{\arraystretch}{0.85}
\begin{adjustbox}{max width=\textwidth}
\begin{tabular}{lccccccc}
\toprule
\multirow{2}{*}{\textbf{Agent}} & \multicolumn{5}{c}{\textbf{RACE}} & \multicolumn{2}{c}{\textbf{FACT}} \\
\cmidrule(lr){2-6} \cmidrule(lr){7-8}
& \textbf{Comp.} & \textbf{Insight} & \textbf{Inst.} & \textbf{Read.}\textsuperscript{\hyperlink{note:read}{$\dagger$}} & \textbf{Overall} & \textbf{Eff. c.} & \textbf{C. acc.}\\ \midrule
doubao-research & 44.84 & 40.56 & 47.95 & 44.69 & 44.34  & 52.62 & 52.86\\
kimi-research  & 44.96 & 41.97 & 47.14 & 45.59 & 44.64  & - & - \\
Claude-research & 45.34 & 42.79 & 47.58 & 44.66 & 45.00 & - & - \\
openai-deepresearch & \textbf{46.46} & 43.73 & \underline{49.39} & \underline{47.22} & 46.45 & 39.79 & \textbf{75.01}  \\
\midrule
WebWeaver (Qwen3-30B-A3B-Instruct-2507)  & 45.15 & \underline{45.78} & 49.21 & \cellcolor{green!10}\textbf{47.34} & \underline{46.77} & 26.74& 25.00 \\
\textbf{DeepWeaver (Qwen3-30B-A3B-Instruct-2507)}  & \cellcolor{green!10}\underline{45.99} & \cellcolor{green!10}\textbf{47.59} & \cellcolor{green!10}\textbf{49.41} & 42.54 & \cellcolor{green!10}\textbf{47.05} & \cellcolor{green!10}\textbf{60.13} & \cellcolor{green!10}\underline{62.02} \\
\bottomrule
\end{tabular}
\end{adjustbox}
}
\vspace{-5pt}
\caption{Results on DeepResearch Bench. DeepWeaver improves comprehensiveness (Comp.) and insight, and achieves substantially better effective citation (Eff. C.) and citation accuracy (C. Acc.) than WebWeaver. Better scores are highlighted in green. All scores are assigned by LLM judges according to detailed expert-level criteria.}
\vspace{-10pt}
\label{tab:drb}
\end{table*}

\subsection{Extension to Web-Based Deep Research}

Web-based deep-research agents also follow a retrieve-then-generate paradigm: search the web, collect evidence, and synthesize long-form answers from open-domain sources. We study whether DeepWeaver can serve as an evidence-weaving module to improve deep-research answer quality.

Since retrieval is not the focus of DeepWeaver, we reuse the web pages and top-level section titles collected by WebWeaver's planner for simplicity and fair comparison\footnote{In terms of cost, DeepWeaver avoids using a summary model to extract webpage-level evidence during evidence collection, substantially reducing the planning-stage cost bottleneck. See \S\ref{sec:cost} for the cost comparison.}. Each question is decomposed into section-level writing tasks. For each section, DeepWeaver treats the collected web pages as evidence fragments and generates the corresponding top-level answer section (\S\ref{sec:drsetting}).

Unlike WebWeaver, which summarizes evidence into coarse-grained outline titles at the webpage level, DeepWeaver operates on fine-grained evidence chunks and weaves them into detailed claim-level structures. As shown in Table~\ref{tab:drb}, DeepWeaver achieves the best overall RACE score. \textbf{It weaves evidence more deeply into insightful and comprehensive claims}. Notably, our method also \textbf{substantially improves effective citation and citation accuracy} over WebWeaver on FACT, indicating that it organizes web evidence into reliable support for answer claims. \textsuperscript{\hypertarget{note:read}{$\dagger$}}However, denser citations and more in-depth claims may affect readability, and the metric may favor an outline-driven writing style. Since readability optimization is not the primary focus of this paper, we leave it to future work.

Despite this trade-off, the overall performance gains confirm that \textbf{evidence weaving also benefits web-based deep research} as a general downstream module beyond knowledge-base QA.

\subsection{Human Evaluation}
\label{sec:he}

To complement the automatic evaluation, we conduct an expert manual assessment on 8 randomly sampled LoQA questions. We recruit three senior PhD students in water-resources engineering to compare DeepWeaver with five representative baselines under blinded conditions, using Win/Tie/Loss judgments along six dimensions: utility, breadth, insight, detail preservation, citation grounding, and overall assessment. We provide detailed description in Appendix \S\ref{sec:anno_dimensions}.

DeepWeaver, RAG, Chain-of-Agents, and STORM \citep{shao2024assisting} use DeepSeek-V4-Pro as their backbone model. HydroGLM is a domain-specific model fine-tuned from GLM-5.0 \citep{glm5team2026glm5vibecodingagentic} on a water-resources and hydropower corpus. The cells in Table~\ref{tab:human_results} report the percentages of Win/Tie/Loss judgments for DeepWeaver against each baseline.

\definecolor{tieyellow}{RGB}{180,125,0}
\newcommand{\yellowtext}[1]{\textcolor{tieyellow}{#1}}

\begin{table*}[t]
\centering
\setlength{\tabcolsep}{8.0pt}
{\fontsize{8.5pt}{11pt}\selectfont
\renewcommand{\arraystretch}{1.0}
\begin{adjustbox}{max width=\textwidth}
\begin{tabular}{lcccccc}
\toprule
\textbf{Baseline (vs. DeepWeaver)} & \textbf{Utility} & \textbf{Breadth} & \textbf{Insight} & \textbf{Detail P.} & \textbf{Citation G.} & \textbf{Overall} \\
\midrule
RAG & \green{79.2} / \yellowtext{12.5} / \redtext{8.3} & \green{70.8} / \yellowtext{12.5} / \redtext{16.7} & \green{91.7} / \yellowtext{8.3} / \redtext{0.0} & \green{75.0} / \yellowtext{12.5} / \redtext{12.5} & \green{70.8} / \yellowtext{29.2} / \redtext{0.0} & \green{83.3} / \yellowtext{12.5} / \redtext{4.2} \\
HydroGLM \tablefootnote{\url{https://hanhai-ai.com}} & \green{66.7} / \yellowtext{25.0} / \redtext{8.3} & \green{70.8} / \yellowtext{20.8} / \redtext{8.3} & \green{95.8} / \yellowtext{4.2} / \redtext{0.0} & \green{83.3} / \yellowtext{16.7} / \redtext{0.0} & \green{83.3} / \yellowtext{16.7} / \redtext{0.0} & \green{75.0} / \yellowtext{20.8} / \redtext{4.2} \\
STORM & \green{54.2} / \yellowtext{33.3} / \redtext{12.5} & \green{8.3} / \yellowtext{8.3} / \redtext{83.3} & \green{79.2} / \yellowtext{8.3} / \redtext{12.5} & \green{50.0} / \yellowtext{25.0} / \redtext{25.0} & \green{25.0} / \yellowtext{70.8} / \redtext{4.2} & \green{50.0} / \yellowtext{37.5} / \redtext{12.5} \\
Chain-of-Agents & \green{95.8} / \yellowtext{4.2} / \redtext{0.0} & \green{95.8} / \yellowtext{4.2} / \redtext{0.0} & \green{100.0} / \yellowtext{0.0} / \redtext{0.0} & \green{100.0} / \yellowtext{0.0} / \redtext{0.0} & \green{83.3} / \yellowtext{16.7} / \redtext{0.0} & \green{100.0} / \yellowtext{0.0} / \redtext{0.0} \\
GPT-5.4 High-Thinking & \green{41.7} / \yellowtext{37.5} / \redtext{20.8} & \green{33.3} / \yellowtext{45.8} / \redtext{20.8} & \green{70.8} / \yellowtext{25.0} / \redtext{4.2} & \green{70.8} / \yellowtext{16.7} / \redtext{12.5} & \green{95.8} / \yellowtext{4.2} / \redtext{0.0} & \green{66.7} / \yellowtext{20.8} / \redtext{12.5} \\
\bottomrule
\end{tabular}
\end{adjustbox}
}
\caption{Results of the blinded human evaluation. Each cell reports the percentage distribution of \green{Win} / \yellowtext{Tie} / \redtext{Loss} judgments for DeepWeaver against the corresponding baseline. Each distribution is based on 24 judgments from three annotators across eight questions.}
\label{tab:human_results}
\vspace{-10pt}
\end{table*}

The expert judgments consistently favor DeepWeaver in utility, insight, detail preservation, citation grounding, and overall assessment. In particular, DeepWeaver achieves a 95.8\% win rate over GPT-5.4 High-Thinking in citation grounding and over HydroGLM in insight. STORM is the only baseline that substantially outperforms DeepWeaver in breadth, possibly reflecting its specialization in producing broad, Wikipedia-style articles. Nevertheless, DeepWeaver remains preferred over STORM in utility, insight, detail preservation, and overall assessment. These results provide direct expert evidence that DeepWeaver's gains reflect improved answer usefulness and evidence synthesis rather than artifacts of LLM-based evaluation, demonstrating the practical value of our system.

We further validate the reliability of the automatic evaluator through a consistency analysis against human annotations and judgments from alternative state-of-the-art models (Appendix~\S\ref{sec:consistency_analysis}). DeepSeek-V3.2 achieves 97\% and 93\% agreement with human labels on AS coverage and DP blank recovery, respectively, while maintaining 85.72--88.90\% agreement with alternative model judges. These results support DeepSeek-V3.2 as a reliable and cost-effective judge for LoQA.

\section{Related Work}
\paragraph{Retrieval-Augmented and Attributed Generation.}
Retrieval-augmented generation is a common way to incorporate external knowledge in question answering \citep{guu2020retrieval,ram2023context,xu2024unsupervisedinformationrefinementtraining}. Sparse vector retrieval \citep{robertson2009probabilistic,wang2023query2doc,schick2023toolformer} retrieves relevant text through term-frequency or keyword matching, while dense vector retrieval \citep{lewis2021retrievalaugmentedgenerationknowledgeintensivenlp,hofstätter2022fidlightefficienteffectiveretrievalaugmented,liu2025llama2vecunsupervisedadaptationlarge} encodes the query and text into vectors. 
Recent advances further improve retrieval through self-reflective retrieval control, graph-structured evidence organization, interleaved retrieval and reasoning, and search-based planning \citep{asai2024self, han2024retrieval, wang2024m, hu2025mcts}.

\paragraph{Long-Context Evidence Integration.}
Larger context windows allow models to include more retrieved evidence, but long-context models may still overlook relevant information in long or noisy inputs \citep{liu2024lost,hsieh2024ruler,modarressi2025nolima}. 
A line of work improves how upstream retrieved evidence is presented to the model. 
Context-refinement methods compress, filter, or restructure retrieved passages before generation, including RECOMP, Chain-of-Note, and LongRefiner \citep{xu2023recomp,yu2024chain,jin2025hierarchical}. 
Divide-and-conquer methods \citep{zhang2024chain,zhou2024llm,guo2025tomleveragingtreeorientedmapreduce} split long contexts into smaller segments and aggregate intermediate outputs. These methods reduce context burden by refining input with coarser evidence representations \citep{jiang2024longllmlingua}, but leave the evidence synthesis gap underexplored.

\paragraph{Long-Form Grounded Writing and Deep-Research Agents.}
Long-form grounded writing has been studied through outline-guided writing and reflection-driven generation, mainly for survey generation \citep{yan2025surveyforgeoutlineheuristicsmemorydriven,skarlinski2024language,shao2024assisting,wu2025superwriterreflectiondrivenlongformgeneration}, but these methods are not primarily designed to answer open-ended questions. 
Citation-oriented work further improves attribution and verifiability of long-form answers \citep{huang2024learning, ye2024effective}. 
Recent deep-research agents extend long-form writing to end-to-end QA systems: one line of work focuses on search timing and search-reasoning interaction \citep{li2025search,jin2025search}, another combines evidence collection with outline-based answer writing \citep{li2025webweaver}, and others improve agent capabilities through agentic supervised fine-tuning and reinforcement learning \citep{wu2026webdancer,zheng2025deepresearcher}.

\section{Conclusion}


We present DeepWeaver, a framework for bridging the evidence synthesis gap in open-ended question answering. Rather than treating retrieved evidence as a flat long-context input, DeepWeaver organizes noisy evidence into Thought Block Chains (TBCs), which connect fine-grained evidence fragments with evidence-grounded woven claims. Through iterative evidence weaving, DeepWeaver identifies overlooked evidence, refines incomplete claims, and improves detail preservation and citation grounding. We also introduce LoQA, a high-density evidence benchmark for evaluating evidence-grounded answer generation. Experiments on knowledge-base QA and web-based deep research show that DeepWeaver consistently improves answer quality across multiple LLM backbones. These findings suggest that effective open-ended QA requires explicit evidence synthesis mechanisms between retrieval and generation, especially under noisy and fragmented evidence.

\section{Limitations}
In this work, we study evidence weaving as a way to advance retrieval-augmented open-ended question answering. While DeepWeaver improves evidence synthesis quality, several limitations remain. First, LoQA focuses on Chinese water-environment questions. Although this domain provides dense technical evidence and complex practical problems, it does not cover all languages, domains, or document types. We therefore evaluate DeepWeaver on DeepResearch Bench as a complementary setting, and future work will further enrich LoQA by expanding its metrics and assessment protocols. Second, DeepWeaver mainly improves the evidence synthesis stage after evidence collection. In the web-based setting, we reuse web pages and section titles collected by WebWeaver’s planner to isolate the effect of evidence weaving. Designing web-search agents better aligned with DeepWeaver remains an open problem. Third, DeepWeaver currently focuses on text-only evidence, while more complete answers may involve figures, charts, images, or structured data. Extending evidence weaving to multimodal and structured evidence remains future work. Finally, although DeepWeaver improves evidence synthesis, model-generated content still requires careful verification.

\section*{Acknowledgements}

This work is supported by the Jing-Jin-Ji Regional Integrated Environmental Improvement--National Science and Technology Major Project (Grant No. 2026ZD1212206), the National Natural Science Foundation of China (Grant No. 52539001), and a grant from the Institute for Guo Qiang, Tsinghua University (Grant No. 2019GQB0003).

\newpage

\bibliography{custom}

\newpage
\appendix

\section{Appendix}
\label{sec:appendix}
\subsection{The LoQA Benchmark}
\label{sec:loqa_benchmark_detail}

\subsubsection{Data Statistics}

As shown in Table~\ref{tab:loqa_stats}, LoQA comprises 100 questions, each accompanied by an average of 206K evidence tokens and 91.79 relevant chunks. 
For evaluation, LoQA includes 2,045 atomic scoring points and 4,779 cloze blanks, corresponding to 20.45 scoring points and 47.79 blanks per question on average. 
The reference answers contain 9,858.64 tokens per question on average. 
These statistics indicate that LoQA is built on extensive noisy evidence and is designed to stress-test evidence synthesis quality, evaluating whether QA systems can comprehensively use evidence, preserve fine-grained domain-specific details, and generate well-cited answers.

\begin{table}[hb]
\centering
\small
\begin{tabular}{l r}
\toprule
\textbf{Statistic} & \textbf{Value} \\
\midrule
\# Questions & 100 \\
Avg. evidence tokens / question & 206,550.41 \\
Avg. relevant chunks / question & 91.79 \\
\midrule
Avg. scoring points / question & 20.45 \\
Avg. cloze blanks / question & 47.79 \\
\midrule
Avg. reference-answer tokens / question & 9,858.64 \\
\bottomrule
\end{tabular}
\caption{\textbf{Statistics of LoQA.} 
Each question is paired with a large, noisy evidence pool of approximately 200K tokens, enabling fine-grained evaluation of evidence synthesis quality.}
\label{tab:loqa_stats}
\end{table}








\subsubsection{Construction Details}
\label{sec:loqa_con}
\begin{figure*}[t]
    \centering
    \includegraphics[width=1.0\linewidth]{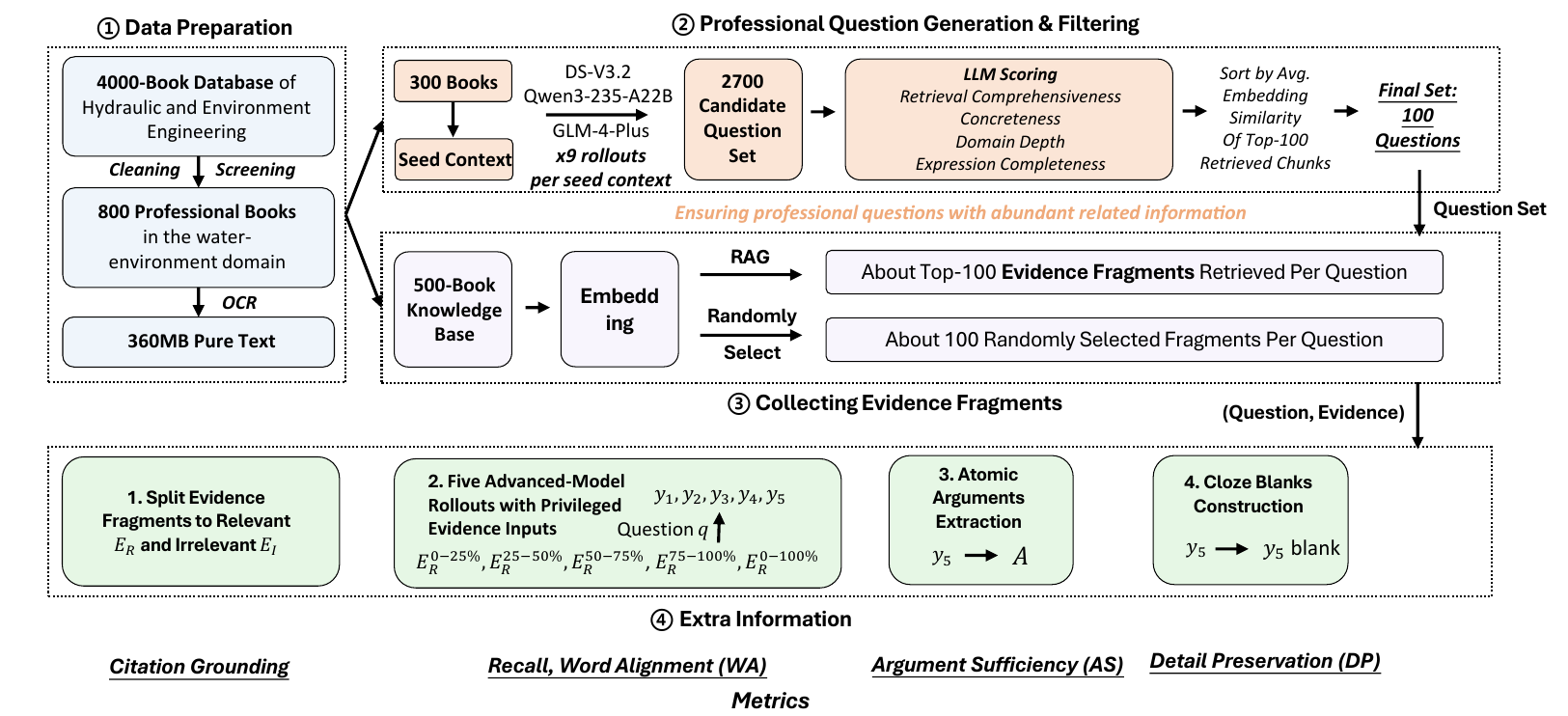} 
    \vspace{-15pt}
    \caption{\textbf{Overview of LoQA construction and evaluation.}
LoQA is built from a water-environment expert knowledge base through data preparation, professional question generation and filtering, and evidence-fragment collection. 
For evaluation, the evidence pool is split into relevant and irrelevant fragments, multiple reference answers are generated from different evidence partitions, and atomic arguments and cloze blanks are constructed to measure citation grounding, content sufficiency, and detail preservation.}
    \label{fig:loqa_con}
    \vspace{-15pt}
\end{figure*}
Figure~\ref{fig:loqa_con} illustrates the construction pipeline of LoQA. 
We construct LoQA from a large-scale expert corpus in the water-environment domain. 
Starting from approximately 30,000 professional PDF books in our hydraulic engineering book list, we first select 4,484 books with relevant tags and then use an LLM to further identify books related to water ecology, water environment, and water circulation based on their titles, keywords, and descriptions. 
This filtering process yields 800 relevant books, from which we randomly select 500 as the knowledge base and use the remaining books as seed documents for practical question generation. 
We extract plain text from the PDFs using PDF-Extract-Kit \citep{wang2024mineru,zhao2024doclayoutyoloenhancingdocumentlayout}, resulting in approximately 360 MB of clean text.

\paragraph{Question generation and filtering.}
To construct professional, open-ended questions, we randomly sample fragments from the seed documents and prompt GLM-4-Plus, DeepSeek-V3.2, and Qwen3-235B-A22B to generate questions with diverse styles and perspectives. 
Each model produces three questions for each seed fragment, yielding 2,700 candidate questions in total. 
We require the questions to be decontextualized, concise, generally answerable, and dependent on multi-source domain knowledge. 
We then use DeepSeek-V3.2 to score each question along four dimensions: retrieval comprehensiveness, concreteness, domain depth, and expression completeness. 
Questions that are vague, overly narrow, insufficiently extensible, or ambiguously expressed are removed. 
We further select the 100 questions with the highest embedding-based cosine similarity to their top-100 retrieved chunks. 
These questions are retained as the final test set. (Figure \ref{fig:loqa_con} \ding{173})

\paragraph{Evidence collection.}
We segment the 500-book knowledge base into 1,024-token fragments and encode each fragment using GLM-Embedding-3. 
For each question, we retrieve approximately 90--100 fragments with the highest cosine similarity to the question embedding:
\[
E_{\mathrm{rag}}
=
\operatorname{TopK}_{e_i \in E}
\left(
\left\langle
\operatorname{emb}(e_i), \operatorname{emb}(q)
\right\rangle
\right)
\]
To simulate realistic noisy retrieval, we additionally sample about 100 random fragments from the corpus. 
As a result, each question is paired with 200 evidence fragments, yielding an evidence pool of approximately 200K tokens. (Figure \ref{fig:loqa_con} \ding{174})
We further use DeepSeek-V3.2 to assess the relevance of each fragment to the question, providing  annotations for citation-grounding evaluation.

\paragraph{Evaluation preparation.}
Because open-ended professional questions typically do not admit a single deterministic gold answer, LoQA constructs multiple possible reference answers synthesized from privileged evidence inputs. 
For each question, we shuffle the relevant evidence fragments $E_R$ and split them into four disjoint partitions, together with the full relevant-evidence set:
\[
E_R^{0\text{-}25\%},\ 
E_R^{25\text{-}50\%},\ 
E_R^{50\text{-}75\%},\ 
E_R^{75\text{-}100\%},\ 
E_R^{0\text{-}100\%}
\]
DeepSeek-V3.2 generates one reference answer from each evidence subset, yielding five reference answers in total. 
The partition-based answers promote local, fine-grained coverage while being generated under a lower context burden, whereas the full-evidence answer provides a global synthesis. 

From these reference answers, we construct more evaluation targets. 
First, we extract atomic arguments from the five reference answers via LLM. 
These arguments capture key evidence-supported answer aspects and are used to assess whether a generated answer covers the major content implied by the question and evidence. 
Second, we construct cloze-style blanks from the full-evidence reference answer. 
These blanks primarily target domain-specific terms, concepts, and short phrases, and are used to assess whether the generated answer preserves fine-grained professional details. 
Together with evidence relevance annotations, these targets enable LoQA to evaluate the answer along three dimensions: citation grounding, content sufficiency, and detail preservation. (Figure \ref{fig:loqa_con} \ding{175})

Although LoQA uses DeepSeek-V3.2 for annotation and verification, it does not rely on holistic subjective scoring of long-form answers. 
Instead, we decompose evaluation into short-context extraction and verification tasks, including evidence relevance judgment, atomic argument matching, and cloze-blank recovery judgment. 
These localized decisions are substantially more constrained than open-ended preference scoring and fall within the reliable capability range of recent advanced LLMs, making DeepSeek-V3.2 suitable for constructing LoQA evaluation signals.

\subsubsection{Comparison with Prior Benchmarks}
\begin{table*}[t]
\centering
\small
\renewcommand{\arraystretch}{1.25}
\setlength{\tabcolsep}{5pt}
\begin{tabular}{p{2.3cm} p{2.8cm} p{3.2cm} p{6.0cm}}
\toprule
\textbf{Benchmark} & \textbf{Task Focus} & \textbf{Evidence Setting} & \textbf{Main Difference from LoQA} \\
\midrule

ALCE \citep{gao2023enabling} 
& Citation-grounded text generation 
& Retrieval corpora built from ASQA, QAMPARI, and ELI5 
& Focuses on fluency, correctness, and citation quality. Its tasks primarily involve producing short answers or entity lists from retrieved passages, whereas LoQA targets comprehensive answers grounded in knowledge-intensive evidence pools of approximately 200K tokens. \\

FreshWiki \citep{shao2024assisting} 
& Wikipedia-like article generation from scratch 
& Web sources collected during the pre-writing stage 
& Evaluates whether systems can research a topic, construct an outline, and generate Wikipedia-like articles. In contrast, LoQA uses a noisy evidence pool and emphasizes comprehensive evidence synthesis in question answering, rather than outline construction or survey writing. \\

LongBench \citep{bai2024longbench} 
& Long-context understanding 
& Single-document QA, multi-document QA, summarization, few-shot learning, synthetic and code tasks  
& Assesses general long-context capability, but most tasks require short-form outputs or automatically verifiable answers rather than open-ended long-form answers. \\

L-Eval \citep{an2024eval} 
& Long-context model evaluation 
& Long documents across diverse domains and task types 
& Covers long-context QA and generation, but many tasks focus on short factual verification, extractive QA, or summarization rather than comprehensive open-ended answering. \\

InfiniteBench \citep{zhang2024bench} 
& Extremely long-context processing 
& 100K+ token contexts across retrieval, code, math, novels, and dialogue 
& Stresses context length and reasoning capacity, but its targets are mainly verifiable short answers instead of long-form citation-grounded answers. \\

Loong \citep{wang2024loong} 
& Extended multi-document QA 
& Multiple relevant documents where missing any document may hurt the answer 
& Emphasizes reasoning over all relevant documents, while LoQA further introduces noisy irrelevant evidence and evaluates citation grounding, content sufficiency, and detail preservation. \\

DeepResearch Bench \citep{du2025deepresearch} 
& End-to-end deep-research agents 
& Web-scale research tasks across diverse fields 
& Evaluates complete deep-research agents, including search and answer generation. LoQA provides a controlled setting to isolate evidence synthesis under a given noisy evidence pool. \\

\textbf{LoQA (Ours)} 
& Professional open-ended answer generation 
& About 200K evidence tokens per question with both relevant and noisy fragments 
& Evaluates evidence synthesis quality, assessing whether QA systems can transform extensive, noisy, and fragmented evidence into comprehensive, well-cited, and detail-preserving answers. \\

\bottomrule
\end{tabular}
\caption{\textbf{Comparison with prior benchmarks.} 
Unlike benchmarks that emphasize short-form answers, survey-style article generation, general long-context understanding, or end-to-end web research, LoQA isolates evidence synthesis quality for open-ended QA by pairing each question with a fixed, extensive, and noisy evidence pool and evaluating citation grounding, content sufficiency, and detail preservation.}
\label{tab:benchmark_comparison}
\vspace{-10pt}
\end{table*}

Table~\ref{tab:benchmark_comparison} compares LoQA with representative benchmarks for long-form QA, citation-grounded generation, long-context understanding, and deep-research agents. 
Existing benchmarks primarily emphasize explanatory QA, verifiable short-form answering, long-context comprehension, or end-to-end web research. 
In contrast, LoQA is designed to isolate the evidence-utilization stage: each professional open-ended question is paired with a fixed, extensive, and noisy evidence pool, and systems are evaluated by their ability to generate comprehensive, well-cited, and detail-preserving answers that directly address the question. 
LoQA is complementary to prior benchmarks and suitable for studying evidence synthesis quality in open-ended question answering.




\subsection{Details of Extending DeepWeaver to DeepResearch Bench}
\label{sec:drsetting}
DeepResearch Bench primarily evaluates end-to-end Open-Ended Deep-Research (OEDR) systems, in which agents search the web, collect evidence, and generate open-ended long-form answers. 
Because DeepWeaver does not include a search module, we study a controlled setting: given the same search system and comparable inference cost, whether DeepWeaver can improve the evidence-to-answer generation stage.

\begin{figure}[htb]
    \centering
    \includegraphics[width=1.0\linewidth]{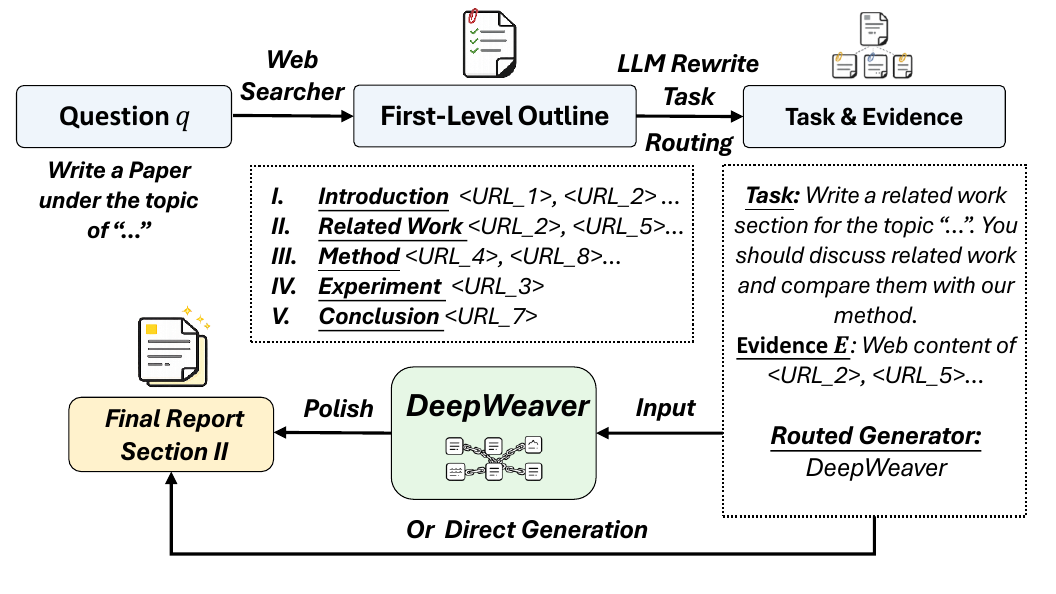} 
    \vspace{-10pt}
    \caption{Overview of DeepWeaver's Extension to Web-Based Deep-Research Scenarios}
    \vspace{-10pt}
\end{figure}

To this end, we use the planner of WebWeaver \citep{li2025webweaver} as the searcher. 
WebWeaver is an advanced dual-agent deep-research framework, where the planner iteratively searches web pages, refines a hierarchical multi-level outline, assigns web pages to outline nodes, and extracts evidence from each page to build a memory bank. 
The writer then fills the outline based on the collected evidence. 
In our setting, however, DeepWeaver does not rely on WebWeaver's detailed outline or extracted evidence memory. 
We only use the first-level outline titles $O_1$ and the corresponding raw web pages $W$ under each title as the input to DeepWeaver. 
This keeps the search layer fixed while allowing us to isolate the effect of evidence weaving during 
long-form answer generation.

For each first-level outline node $o \in O_1$, we first rewrite $o$ into a section-level DeepWeaver task $q$. 
The model also judges (routes) whether the section is critical to answering the original question or contains extensive webpage information. 
If so, we run DeepWeaver to generate the section; otherwise, we use direct generation. 
This rewriting and routing step only consumes a few hundred tokens and has negligible cost. 
For evidence construction, we truncate the raw webpage content associated with $o$ to at most 35K tokens and split it into 1,024-token fragments, which are then regarded as the evidence pool for DeepWeaver.

\begin{table}[htb]
\centering
\small
\begin{tabular}{l r}
\toprule
\textbf{Statistic} & \textbf{Value} \\
\midrule
Avg. first-level outline node $|O_1|$ / question & 4.41 \\
Avg. routed outline node / question & 2.35 \\
Avg. routed-item ratio / question & 59.39\% \\
\midrule
Avg. evidence fragments / item & 88.92 \\
Avg. evidence fragments / routed item & 98.00 \\
Avg. raw tokens of evidence / routed item & 92,752.79 \\
\bottomrule
\end{tabular}
\caption{\textbf{Routing statistics of DeepWeaver on DeepResearch Bench.} 
Each question contains 4.41 first-level outline nodes on average, among which 2.35 nodes are routed to DeepWeaver, accounting for 59.39\% of the items. Routed items are evidence-heavy, containing 98 chunks and 92.75K raw tokens on average. In LoQA, the evidence length is much larger, with about 200K tokens per question, indicating that \textbf{DeepWeaver is sufficient to handle evidence-context lengths in realistic applications, such as multi-webpage reading.} }
\label{tab:routing_stats}
\vspace{-10pt}
\end{table}

After DeepWeaver generates the section-level answers, we apply a lightweight polishing stage to improve fluency and readability. 
We then add an introduction and a conclusion to produce a complete end-to-end answer. 
These polishing, introduction, and conclusion stages are auxiliary components introduced solely to match the answer format required by DeepResearch Bench. The core comparison focuses on whether DeepWeaver can more effectively synthesize web evidence into comprehensive, insightful, and well-cited answer content.

\subsection{Cost Analysis}

\label{sec:cost}
We analyze the cost of DeepWeaver on DeepResearch Bench by reusing WebWeaver's Searcher (Planner) and replacing the downstream writer with DeepWeaver. To reduce variance introduced by API providers, network latency, and webpage accessibility, we exclude the cost and runtime of web search and page access, and measure only LLM-side computation. Local runtime is measured with Qwen3-30B-A3B-Instruct-2507 services on local $2\times$A100 GPUs.

As shown in Table~\ref{tab:cost}, DeepWeaver incurs a monetary cost comparable to WebWeaver (\$0.313 vs. \$0.303), while reducing local inference time from 18.5 to 14.2 minutes. Although DeepWeaver consumes more input tokens (3.16M vs. 2.35M), it substantially reduces output tokens from 307K to 94K by avoiding explicit webpage-level evidence extraction and memory-bank construction. This extraction-free design reallocates the budget from generating intermediate summaries to reading more original evidence. 

\begin{figure*}[h]
    \centering
    \includegraphics[width=0.95\linewidth]{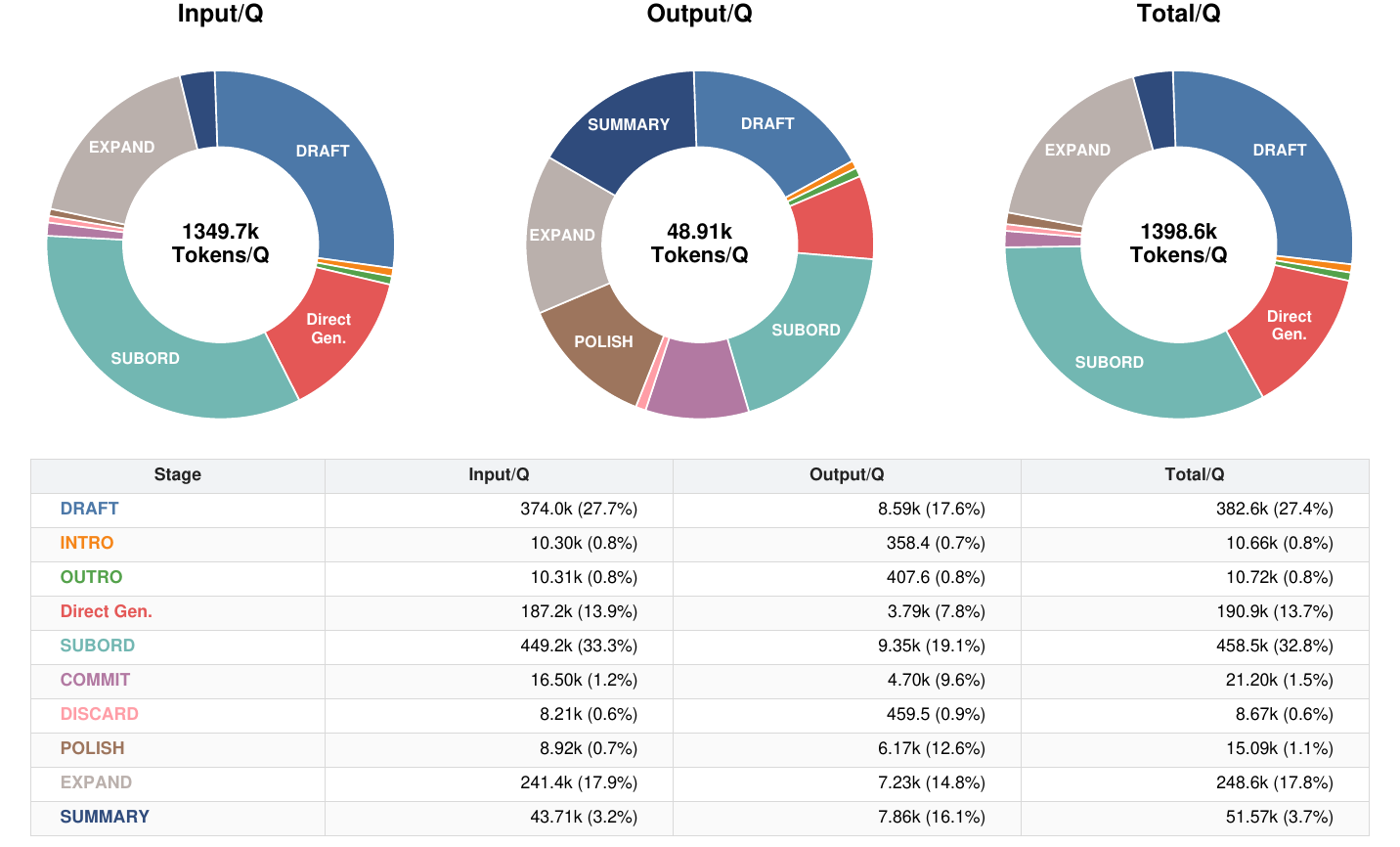} 
    \vspace{-5pt}
    \caption{\textbf{Token usage of DeepWeaver (excluding the searcher) across stages.} 
We report token consumption on DeepResearch Bench. 
POLISH, INTRO, and OUTRO denote polishing, introduction generation, and conclusion generation, respectively (\S \ref{sec:drsetting}). 
Although the \textit{Subordinate} stage is the main additional cost, its output tokens are only slightly higher than the \textit{Summary} output, and its total tokens are comparable to the \textit{Draft stage}, showing that TBC-based evidence weaving is lightweight beyond evidence input and answer generation.}
    \label{fig:token_usage}
    \vspace{-15pt}
\end{figure*}

Under comparable cost, DeepWeaver can pass up to 35K tokens of original content per webpage to the writer, compared with 24K tokens in WebWeaver, reducing the cost per observed token from $1.26\times10^{-5}$ to $8.94\times10^{-6}$. Figure~\ref{fig:token_usage} further shows that DeepWeaver introduces only lightweight internal overhead. The main additional cost comes from the \textit{Subordinate} stage, which inspects residual evidence and enriches the TBC. However, its output-token usage is only slightly higher than that of the \textit{Draft} and \textit{Summary} stages, and its total token usage remains comparable to the \textit{Draft} stage. Other maintenance stages, including \textit{Commit}, \textit{Discard}, \textit{Polish}, \textit{Intro}, and \textit{Outro}, account for only a small fraction of the total cost. Overall, DeepWeaver improves evidence utilization not by substantially increasing cost, but by reallocating the same budget: it avoids expensive intermediate output generation and explicit evidence extraction, and instead uses lightweight TBC metadata to route more original evidence into focused answer generation.

\newcommand{\cmark}{\textcolor{green!60!black}{\ding{51}}} 
\newcommand{\xmark}{\textcolor{red}{\ding{55}}}    
\begin{table*}[h]
\centering
\setlength{\tabcolsep}{8.0pt}
{\fontsize{10pt}{14pt}\selectfont
\renewcommand{\arraystretch}{1.0}
\begin{adjustbox}{max width=\textwidth}
\begin{tabular}{lcccccc}
\toprule
\textbf{Agent}
& \makecell{\textbf{Evidence-}\\ \textbf{Extraction-Free*}} & \makecell{\textbf{\# Input}\\ \textbf{Tokens}}
& \makecell{\textbf{\# Output} \\ \textbf{Tokens}} & \makecell{\textbf{Local}\\ \textbf{Time (min)}} & \textbf{Cost} & \makecell{\textbf{Answer}\\ \textbf{Len.}}\\ \midrule
WebWeaver & \xmark & \textbf{2,348K} & 307K & 18.5  & \textbf{\$0.303} & 11,889.3 \\
\textbf{DeepWeaver} & \cmark & 3,161K & \textbf{94K} & \textbf{14.2} & \$0.313 & 12,303.8\\
\midrule
\textbf{Agent} & \makecell{\textbf{\# Tokens Seen by Writer}\\/  \textbf{Web Page}} & \multicolumn{2}{c}{\makecell{\textbf{\# Input Tokens}\\/\textbf{Token Seen}}} & \multicolumn{2}{c}{\makecell{\textbf{\# Output Tokens}\\/\textbf{Token Seen}}} &  \makecell{\textbf{Cost}\\/\textbf{Token Seen} } \\ \midrule
WebWeaver & 24,000 &  \multicolumn{2}{c}{97.83} & \multicolumn{2}{c}{12.79} & \$$1.26\times 10^{-5}$ \\
\textbf{DeepWeaver}  & \textbf{35,000} \textsuperscript{$\dagger$} & \multicolumn{2}{c}{\textbf{90.31}} & \multicolumn{2}{c}{\textbf{2.69}} & \$\textbf{$8.94\times 10^{-6}$}\\
\bottomrule
\end{tabular}
\end{adjustbox}
}
\vspace{-5pt}
\caption{\textbf{Average cost comparison between WebWeaver and DeepWeaver (including the searcher).} 
Under comparable cost and with the same searcher, replacing the downstream writer with DeepWeaver improves DeepResearch Bench performance. 
To reduce evaluation variance caused by API providers, network latency, and web access, we exclude the cost and time of web search and page access. 
Local Time only measures inference time of Qwen3-30B-A3B-Instruct-2507 on local $2\times$A100 GPUs. 
\textbf{*} DeepWeaver avoids extracting evidence content and maintaining a webpage-level memory bank, substantially reducing output-token overhead. 
\textsuperscript{$\dagger$} Under comparable cost, DeepWeaver can pass more original webpage content to the answer writer, up to 35K tokens, whereas WebWeaver relies on evidence extracted from 24K-token webpage inputs.}
\vspace{-5pt}
\label{tab:cost}
\end{table*}

\subsection{Consistency Analysis}
\label{sec:consistency_analysis}
To assess the reliability of the LLM-based evaluation, we audit the automatic judgments used for Argument Sufficiency (AS) and Detail Preservation (DP) against human annotations and judgments from alternative state-of-the-art models. The audit examines both argument extraction and the subsequent LLM judger decisions.

\paragraph{Argument extraction.}
We randomly sample 100 extracted arguments by the judger, DeepSeek-V3.2, and assess whether each extraction is semantically reasonable and relevant to the corresponding question. As shown in Table~\ref{tab:as_extraction_agreement}, human annotators consider all 100 extractions reasonable and 97 relevant to the question. GPT-5.2 and DeepSeek-V4-Pro produce similar judgments, indicating that argument extraction in AS yields well-formed and predominantly question-relevant evaluation targets.

\begin{table}[htbp]
\centering
\setlength{\tabcolsep}{4.8pt}
{\fontsize{9pt}{10pt}\selectfont
\renewcommand{\arraystretch}{0.95}
\begin{adjustbox}{max width=\columnwidth}
\begin{tabular}{lcc}
\toprule
\textbf{Annotator} & \makecell{\textbf{Reasonable}\\\textbf{Extraction}} & \makecell{\textbf{Question}\\\textbf{Relevance}} \\
\midrule
Human & 100 / 100 & 97 / 100 \\
GPT-5.2 & 99 / 100 & 92 / 100 \\
DeepSeek-V4-Pro & 100 / 100 & 99 / 100 \\
\bottomrule
\end{tabular}
\end{adjustbox}
}
\caption{Agreement evaluation of argument extraction for Argument Sufficiency (AS).}
\label{tab:as_extraction_agreement}
\end{table}

\paragraph{AS coverage judgments.}
We further audit the binary decisions of whether a generated answer covers each extracted argument. Table~\ref{tab:as_coverage_agreement} reports pairwise raw agreement. The results show that DeepSeek-V3.2 agrees with human annotations on 97 of 100 audited cases. Across all 2,045 AS arguments, it achieves 88.90\% agreement with GPT-5.2 and 85.72\% with DeepSeek-V4-Pro.

\begin{table}[htbp]
\centering
\setlength{\tabcolsep}{4.8pt}
{\fontsize{9pt}{10pt}\selectfont
\renewcommand{\arraystretch}{0.95}
\begin{adjustbox}{max width=\columnwidth}
\begin{tabular}{lc}
\toprule
\textbf{Annotator} & \textbf{Agreement (\%)}  \\
\midrule
Human & 97 / 100 (97.00\%)  \\
GPT-5.2 & 1,818 / 2,045 (88.90\%)  \\
DeepSeek-V4-Pro & 1,753 / 2,045 (85.72\%)  \\
\bottomrule
\end{tabular}
\end{adjustbox}
}
\caption{Agreement evaluation of AS coverage judgments across human and model annotators.}
\label{tab:as_coverage_agreement}
\end{table}

\paragraph{DP blank-recovery judgments.}
DP evaluates whether fine-grained information removed from a reference answer can be recovered from the generated answer. As reported in Table~\ref{tab:dp_recovery_agreement}, DeepSeek-V3.2 agrees with human labels on 93 of 100 sampled blank-recovery cases. Across 4,649 blanks, it achieves 86.62\% agreement with GPT-5.2. These results indicate that DeepSeek-V3.2 provides a reliable and cost-effective evaluator for automatic assessment.

\begin{table}[htbp]
\centering
\setlength{\tabcolsep}{4.8pt}
{\fontsize{9pt}{10pt}\selectfont
\renewcommand{\arraystretch}{0.95}
\begin{adjustbox}{max width=\columnwidth}
\begin{tabular}{lc}
\toprule
\textbf{Annotator} & \textbf{Agreement} \\
\midrule
Human & 93/100 (93.00\%) \\
GPT-5.2 & 4,027/4,649 (86.62\%) \\
\bottomrule
\end{tabular}
\end{adjustbox}
}
\caption{Agreement audit of Detail Preservation (DP) blank-recovery judgments.}
\label{tab:dp_recovery_agreement}
\end{table}

Overall, the human audits show high agreement with DeepSeek-V3.2 for both AS coverage (97\%) and DP blank recovery (93\%), while the cross-model comparisons remain broadly consistent despite differences in model-specific decision thresholds. These results support the design of LoQA, which decomposes long-form answer evaluation into localized binary sub-tasks rather than relying on a single holistic score. Considering both empirical agreement and evaluation cost, we retain DeepSeek-V3.2 as the default validated evaluator for AS and DP.

\subsection{Human Validation Protocol}
\label{sec:anno_dimensions}

This section details the human evaluation protocol described in \S\ref{sec:he}. Each annotator was compensated 15 RMB per evaluated question. The study collected no personally identifiable or sensitive information and posed no apparent ethical risks. The evaluation criteria for the six dimensions are provided in Table~\ref{tab:annotation_dimensions}.

\begin{table*}[t]
  \centering
  \setlength{\tabcolsep}{6.0pt}
  {\fontsize{8.8pt}{12pt}\selectfont
  \renewcommand{\arraystretch}{1.12}
  \begin{adjustbox}{max width=\textwidth}
  \begin{tabular}{p{0.18\textwidth} p{0.76\textwidth}}
  \toprule
  \textbf{Dimension} & \textbf{Instruction} \\
  \midrule

  \textbf{Utility} &
  Judge whether the answer stays focused on the user question and is practically useful for solving it.

  \textbf{Key criteria:}
  (1) Directly addresses the core question;
  (2) analyzes the central issue instead of mainly presenting generic background;
  (3) provides useful and actionable perspectives, frameworks, or mechanism explanations;
  (4) helps experts understand the problem, broaden thinking, organize information, or form judgments.

  \textbf{Penalties:}
  (1) Excessive unnecessary information with limited response to the question;
  (2) superficially related but peripheral content;
  (3) remotely related content not needed for solving the problem.

  \textbf{Preference rule:}
  Prefer the answer that better captures the question focus and provides more substantive value for expert reasoning or judgment. Choose a tie
  when both are basically responsive and differ little in practical usefulness.
  \\
  \midrule

  \textbf{Breadth} &
  Judge whether the answer sufficiently covers the key perspectives and core arguments needed for the question. Penalize outline-like or
  survey-like writing when most sections do not address the question core or connect directly to it.

  \textbf{Key criteria:}
  (1) Provides multiple perspectives and solutions rather than a single or trivial angle;
  (2) analyzes from several angles, such as mechanisms or technical pathways;
  (3) helps experts build a more complete thinking framework;
  (4) explains how listed ideas logically relate to the question instead of merely piling up aspects.

  \textbf{Penalties:}
  (1) Omits important perspectives or arguments;
  (2) is long but repetitive;
  (3) piles up keywords or items with shallow analysis;
  (4) adds weakly related content, creating pseudo-breadth;
  (5) lists aspects without explaining how they support the answer.

  \textbf{Preference rule:}
  Prefer the answer that naturally covers the required key perspectives and core arguments with substantive analytical value. Choose a tie when
  coverage is similar or differences mainly reflect length rather than substantive coverage quality.
  \\

  \bottomrule
  \end{tabular}
  \end{adjustbox}
  }
  \vspace{-5pt}
  \end{table*}

  \begin{table*}[t]
  \centering
  \setlength{\tabcolsep}{6.0pt}
  {\fontsize{8.8pt}{12pt}\selectfont
  \renewcommand{\arraystretch}{1.12}
  \begin{adjustbox}{max width=\textwidth}
  \begin{tabular}{p{0.18\textwidth} p{0.76\textwidth}}
  \toprule
  \textbf{Dimension} & \textbf{Instruction} \\
  \midrule

  \textbf{Insight} &
  Judge whether the answer provides a deeper, more systematic understanding of the question and offers inspiration for answering it.

  \textbf{Key criteria:}
  (1) Systematically explains and analyzes the problem;
  (2) proposes deeper solutions or original, insightful viewpoints;
  (3) helps experts form higher-level understanding;
  (4) synthesizes fragmented information deeply and uses it to support arguments.

  \textbf{Penalties:}
  (1) Merely lists concepts, terms, or common-sense points without depth;
  (2) lacks explanations of reasons, mechanisms, or evidence behind conclusions;
  (3) sounds professional but relies mainly on jargon or boilerplate, without substantive insight;
  (4) discusses multiple aspects in parallel without deeper connections.

  \textbf{Preference rule:}
  Prefer the answer with clearly deeper understanding and response that better supports expert judgment. Choose a tie when both stay at a
  general analytical level or differ little in insight.
  \\
  \midrule

  \textbf{Detail Preservation} &
  Judge whether the answer preserves and organizes valuable concrete details and professional information, and uses them to support its
  arguments.

  \textbf{Key criteria:}
  (1) Includes professional terms and key information useful for understanding the problem;
  (2) preserves fine-grained information such as mechanisms, steps, technical pathways, constraints, or applicable scenarios;
  (3) uses concrete examples to improve clarity and usability;
  (4) provides specific content beyond high-level analysis to support arguments;
  (5) organizes useful details instead of indiscriminately piling up information.

  \textbf{Penalties:}
  (1) Is too generic or abstract;
  (2) omits key supporting details for important arguments;
  (3) includes many details that are weakly related or of limited help;
  (4) mentions details without effective organization or explanation.

  \textbf{Preference rule:}
  Prefer the answer that preserves more valuable, relevant, and problem-supporting details. Choose a tie when the preservation and usefulness
  of key details are similar.
  \\
  \midrule

  \textbf{Citation Grounding} &
  Judge whether key facts, judgments, and conclusions are sufficiently and accurately supported by cited evidence.

  \textbf{Key criteria:}
  (1) Provides citations near important facts or judgments;
  (2) uses citations that directly or strongly support the corresponding claims;
  (3) cites sufficiently rich and relevant materials;
  (4) makes reasonable generalizations from evidence rather than extending beyond it.

  \textbf{Penalties:}
  (1) Key facts, judgments, or conclusions lack citation support;
  (2) citations are too sparse to support the arguments;
  (3) cited materials are too general, marginally relevant, or weakly related to the question.

  \textbf{Preference rule:}
  Prefer the answer with more sufficient, accurate, and relevant citation support for key arguments. Choose a tie when citation quantity and
  quality are similar, or differences do not affect core credibility.
  \\
  \midrule

  \textbf{Overall Preference} &
  Judge which answer has higher overall quality and better meets experts' needs for understanding and answering the question.

  Consider relevance, breadth, insight, detail preservation, citation grounding, and practical usefulness.

  \textbf{Additional note:}
  Because this study focuses on content quality, prioritize the content itself across all dimensions. Surface differences such as length or
  presentation format should not be the main basis for judgment. Suggestions on presentation or other useful comments may be written in the
  comment box.
  \\

  \bottomrule
  \end{tabular}
  \end{adjustbox}
  }
  \vspace{-5pt}
  \caption{Annotation instructions for the six evaluation dimensions in human expert validation.}
  \vspace{-5pt}
  \label{tab:annotation_dimensions}
  \end{table*}

\subsection{Prompt Templates}

This section presents the English prompt templates used for each stage of DeepWeaver.

\onecolumn

\setpromptboxtitle{Thought Block Chain Structure Example in Original Source Code}
\begin{promptboxlist}
{
    \promptcomment|// Thought Block Chains b_i@
    "logic_blocks": [  
        {
          \promptcomment|// Claims c_i@
          "argument": "Netflix's adaptation of 'One Hundred Years of Solitude' succeeded by prioritizing cultural and linguistic authenticity through filming in Spanish and in Colombia.",
          
          \promptcomment|// Keywords k_i@
          "keywords": [
                "cultural authenticity",
                "original language",
                "Colombian production"
            ],
            
          \promptcomment|// Salient Information s_i@
          "key_info": "The series was filmed in Spanish and exclusively in Colombia, the author's native country, to preserve the novel's rhythm and regional accent, and to draw from the nation's real geography and history. This decision was made to honor the author's conditions and to avoid the 'white-washed Hollywood' approach that had previously failed other adaptations.", 
          
          \promptcomment|// Evidence Fragments E_i@
          "reference_chunks": [5,22,44,82,119],
            
          \promptcomment|// Covered Evidence Fragments, Top-k Embedding Similarity Chunks to c_i+s_i@
          \promptcomment|// Available only for main TBCs@
          "retrieval_reference_chunks": [19,27,15,16,120]
        }, ...
    ]
}
\end{promptboxlist}

\newpage

\newcommand{\imend}{\textbf{\ttfamily\textless|im\_end|\textgreater}}
\newcommand{\imstartsystem}{\textbf{\ttfamily\textless|im\_start|\textgreater system}}
\newcommand{\imstartuser}{\textbf{\ttfamily\textless|im\_start|\textgreater user}}
\newcommand{\imsep}{\imend\imstart}
\newcommand{\promptsep}{%
  \par\vspace{2pt}%
  \noindent{\color{gray!55}\rule{\linewidth}{0.4pt}}%
  \par\vspace{2pt}%
}

\setpromptboxtitle{Draft Stage Prompt Template}
\begin{promptboxlist}
\promptcomment|//Step 1: Draft Answer Generation Prompt Template@
\imstartsystem
You are a professional research assistant and need to answer the question based on the following document passages.
\imend
\imstartuser

Question: \promptcomment|{question}@

Chunk #i: \promptcomment|{evidence fragments}@

Please generate a comprehensive answer report based on the following document passages. If you cite any passage, place the cited passage number in [] at the end of the corresponding sentence, such as [1,3] or [2]:
\imend

\promptsep

\promptcomment|//Step 2: Extracting TBC Prompt Template@
\imstartsystem
You are a professional logic-organization expert. Your task is to break the "generated answer" into a set of logic blocks.

Please output JSON according to the following requirements:
1. Each logic block should revolve around a clear subtopic and reflect the internal logical structure of the answer.
2. Each logic block should contain:
   - argument: a one-sentence summary of the logic block's argument
   - keywords: 3-5 keywords
   - key_info: a summary of the key information corresponding to the argument
   - reference_chunks: the reference passages involved in this argument
3. The reference_chunks indices may be parsed from the displayed citations in the "generated answer", and may also be expanded with other highly relevant reference passages. Each argument should have 3-8 reference_chunks
4. Ensure that the JSON format is correct and can be parsed.

Output format:
{
  "logic_blocks": [
    {
      "argument": "...",
      "keywords": ["...", "..."],
      "key_info": "...",
      "reference_chunks": [1,2,3]
    },
    
    {
      "argument": "...",
      "keywords": ["...", "..."],
      "key_info": "...",
      "reference_chunks": [6,9]
    },
    ...
  ]
}\imend

\imstartuser
Please generate a structured JSON response based on the following references and question:
Question: \promptcomment|{question}@

Chunk #i: \promptcomment|{evidence fragments}@

Generated answer:
\promptcomment|{answer}@
\imend
\end{promptboxlist}

\setpromptboxtitle{Subordinate Stage Prompt Template}
\begin{promptboxlist}
  \promptcomment|//Step 1: Supplementary Answer Generation Prompt Template@
  \imstartsystem
  You are a professional research assistant and need to answer the question based on the following document passages.
  \imend
  \imstartuser

  Chunk #i: \promptcomment|{uncovered evidence fragments}@

  Question: \promptcomment|{question}@

  Below are the logic blocks that have already been identified as discarded for this question:
  \promptcomment|{discarded logic blocks}@

  Please generate a comprehensive answer report based on the following document passages. If you cite any passage, place the cited passage number in [] at the end of the corresponding sentence, such as [1,3] or
  [2]:
  \imend
  \promptsep
  \promptcomment|//Step 2: Extracting Supplementary TBC Prompt Template@
  \imstartsystem
  You are a professional logic-organization expert. Your task is to break the "generated answer" into a set of supplementary logic blocks.

  Please output JSON according to the following requirements:
  1. Each logic block should revolve around a clear subtopic and highlight its supplementary value to the existing logic.
  2. Each logic block should contain:
     - argument: a one-sentence summary of the logic block's argument
     - keywords: 3-5 keywords
     - key_info: a summary of the key information corresponding to the argument
     - reference_chunks: the reference passages involved in this argument
  3. The reference_chunks indices may be parsed from the displayed citations in the "generated answer", and may also be expanded with other highly relevant reference passages. Each argument should have 3-8
  reference_chunks
  4. Ensure that the JSON format is correct and can be parsed.

  Output format:
  {
    "logic_blocks": [
      {
        "argument": "...",
        "keywords": ["...", "..."],
        "key_info": "...",
        "reference_chunks": [1,2,3]
      },

      {
        "argument": "...",
        "keywords": ["...", "..."],
        "key_info": "...",
        "reference_chunks": [6,9]
      },
      ...
    ]
  }
  \imend

  \imstartuser
  Please generate a structured JSON response based on the following references and question:

  Chunk #i: \promptcomment|{uncovered evidence fragments}@

  Question: \promptcomment|{question}@

  Below are the logic blocks that have already been identified as discarded for this question:
  \promptcomment|{discarded thought blocks}@

  Generated answer:
  \promptcomment|{answer}@

  JSON response:
  \imend
  \end{promptboxlist}

\setpromptboxtitle{Commit Stage, Merge Operation Prompt Template}
\begin{promptboxlist}
  \promptcomment|//Step 1: Merge Candidate Screening Prompt Template@
  \imstartsystem
  You are a professional logic-comparison expert. Your task is to find only the pairs of logic blocks between the main and subord groups that must be merged.

  Requirements:
  1. Consider only cross-group pairings, and do not consider main-main or subord-subord.
  2. Suggest a merge only when the viewpoints of two logic blocks clearly overlap; otherwise do not output it.
  3. The reference_blocks in each suggestion must be a pair, and it must consist of exactly one positive main index and one negative subord index.
  4. What you output are candidate suggestions, not the final result; when in doubt, leave it out.
  5. If there is no candidate pair, please return an empty array [].
  6. Output at most \promptcomment|{k}@ of the most important overlaps.

  Output format:
  [
    {
      "reference_blocks": [2, -1],
      "rationale": "Both discuss ......, and the difference between them is ..., described in a neutral tone",
      "overlap_ratio": 0.xx
    }
  ]
  \imend

  \imstartuser
  Please screen the must-merge candidate pairs according to the following main and subord logic block information:

  Please organize and merge the main and subord logic blocks based on the following information:

  **Question:**
  \promptcomment|{question}@

  **Main logic blocks (numbered 1, 2, 3, ...):**
  **Main Logic Block i:**
  Argument: \promptcomment|{main argument}@
  Keywords: \promptcomment|{main keywords}@
  Key information: \promptcomment|{main key information}@

  **Subord logic blocks (numbered -1, -2, -3, ...):**
  **Subord Logic Block -i:**
  Argument: \promptcomment|{subord argument}@
  Keywords: \promptcomment|{subord keywords}@
  Key information: \promptcomment|{subord key information}@

  Notes:
  - Do not output a single index.
  - Do not output one-to-many or many-to-one.
  - If there is no candidate pair that meets the criteria, return [] directly.

  JSON response:
  \imend

  \promptsep

  \promptcomment|//Step 2: Commit Thought Block Merging Prompt Template@
  \imstartsystem
  You are a professional expert in logical organization and information integration. Your task is to organize and merge logic blocks based on the provided main and subord logic block information, forming a
  clearer and more complete logical structure.

  Please analyze and merge the logic blocks according to the following requirements:
  1. **Input format**: There are two lists of logic blocks: the main logic block list and the subord logic block list.
  2. **Identify repeated or similar arguments**: Analyze the main and subord logic blocks and identify pairs of logic blocks with repeated or similar arguments. It is already known that there are no repeated
  viewpoints within main or within subord, so you only need to consider whether there are two viewpoints across groups that are highly similar or repeated.
  3. **Organize the logical order reasonably**: Organize the logic blocks in a reasonable order to form merged logic blocks.
  4. **Extract merged information**: For each merged logic block, extract the merged argument, keywords, and the indices of the original logic blocks it cites. Main indices are 1,2,3,..., and subord indices are
  -1,-2,-3,....
  5. **Citation format and requirements**: reference_blocks can only be a single number or a **pair**, and a pair **must** consist of one main logic block (positive index) and one subord logic block (negative
  index).
  6. **Overall requirements**: Ensure that every logic block is covered, **and ensure that no two merged logic blocks contain the same reference_blocks.** Each index may appear only once in all reference_blocks
  dicts.

  Output format requirements:
  Please output in JSON format, as shown below:

  [
      {
          "reference_blocks": 1,
          "argument": "Merged argument 1",
          "keywords": ["Keyword 1", "Keyword 2"]
      },
      {
          "reference_blocks": [-1, 2],
          "argument": "Merged argument 2",
          "keywords": ["Keyword 3", "Keyword 4"]
      },
      {
          "reference_blocks": 3,
          "argument": "Merged argument 3",
          "keywords": ["Keyword 5", "Keyword 6"]
      },
      {
          "reference_blocks": [-2, 4],
          "argument": "Merged argument 4",
          "keywords": ["Keyword 6", ...]
      },
      ...
  ]
  \imend

  \imstartuser
  Please organize and merge the main and subord logic blocks based on the following information:

  **Question:**
  \promptcomment|{question}@

  **Main logic blocks (numbered 1, 2, 3, ...):**
  **Main Logic Block i:**
  Argument: \promptcomment|{main argument}@
  Keywords: \promptcomment|{main keywords}@
  Key information: \promptcomment|{main key information}@

  **Subord logic blocks (numbered -1, -2, -3, ...):**
  **Subord Logic Block -i:**
  Argument: \promptcomment|{subord argument}@
  Keywords: \promptcomment|{subord keywords}@
  Key information: \promptcomment|{subord key information}@

  Below are the pre-generated merge candidate suggestions:
  \promptcomment|{candidate merge pairs and rationales}@

  Notes:
  - Ensure that the JSON format is correct and can be parsed.
  - Keep the logic clear and well-structured. Each merged argument should be concise and clear rather than verbose. If it is verbose, it must be split into two arguments and described separately rather than
  merged into one logic block.
  - Ensure that every logic block is covered, **and ensure that no two logic blocks contain the same reference_blocks.** Each index may appear only once in all reference_blocks dicts.

  JSON response:
  \imend
  \end{promptboxlist}

\setpromptboxtitle{Commit Stage, Discard Operation Prompt Template}
\begin{promptboxlist}
  \promptcomment|//Thought Block Discard Prompt Template@
  \imstartsystem
  You are a professional logic-analysis expert. Your task is to analyze logic blocks and identify the logic blocks that should be discarded.

  Please analyze the logic blocks according to the following criteria:

  1. Off-topic: logic blocks unrelated to the theme of the question.
  2. Redundant information: logic blocks whose content is obviously consistent with, or seriously repetitive of, other existing logic blocks.
  3. Mixed incorrect information: logic blocks containing obvious errors or information inconsistent with facts.
  4. Low-impact information: peripheral information that does not directly help solve the question.

  Output format requirements:
  Please output in JSON format, containing the two fields block_id and rationale. The rationale field should provide a brief reason for discarding the logic block.
  If no logic block needs to be discarded, please return an empty array [].

  Notes:
  - Return only the logic blocks that need to be discarded.
  - Please judge the degree of discarding carefully. For example, if the question includes expressions such as "as comprehensive as possible", then fewer logic blocks should be discarded, or none at all.
  - Ensure that the JSON format is correct and can be parsed.
  \imend

  \imstartuser
  Please generate a structured JSON response according to the following information:

  Please analyze the following logic blocks and identify which ones should be discarded:

  **Question:**
  \promptcomment|{question}@

  **Logic Blocks:**
  **Logic Block i:**
  Argument: \promptcomment|{argument}@
  Keywords: \promptcomment|{keywords}@
  Key information: \promptcomment|{key information}@

  **Discard Criteria:**
  1. Off-topic: logic blocks that are unrelated to the question.
  2. Redundant information: logic blocks that repeat or closely overlap with other logic blocks.
  3. Mixed or incorrect information: logic blocks that contain clearly incorrect information or factual errors.
  4. Weak-impact information: peripheral information that does not directly help solve the question.

  **Task:**
  Identify all logic blocks that should be discarded, and provide a one-sentence rationale for each discarded block.

  **Output Format:**
  Return the result in JSON format, following this example:

  [
      {
          "block_id": 1,
          "rationale": "This logic block is unrelated to the question and therefore off-topic."
      },
      {
          "block_id": 3,
          "rationale": "This logic block overlaps with logic block 2 and is therefore redundant."
      }
  ]

  If no logic blocks need to be discarded, return an empty array `[]`.

  JSON response:
  \imend
  \end{promptboxlist}

\setpromptboxtitle{Expand Stage Prompt Template}
\begin{promptboxlist}
  \promptcomment|//Logic Block Expansion Prompt Template@
  \imstartsystem
  You are a professional report-writing expert. Your task is to discuss the argument of a specified logic block among a series of logic blocks, based on the provided literature passages and while staying closely focused on the question itself.

  Requirements:
  1. The discussion must stay closely aligned with the argument and the question, while also maintaining coherence with the already generated preceding answer.
  2. Refer to and cite information from the provided chunk pool comprehensively and extensively.
  3. Use concise and clear language.
  4. Maintain objectivity and accuracy.
  5. Staying close to the question is the first priority. Remove material that is only loosely related to the question.
  6. When a sentence uses information from the chunk pool, cite the local chunk ids at the end of the sentence in the format [1] or [1,2].
  7. Use only the chunk ids that appear in the prompt. Do not invent citation ids.

  Please generate a complete discussion and output the content directly with inline citations.
  \imend

  \imstartuser
  Please write a detailed discussion for the current logic block while keeping coherence with the overall answer.

  Question:
  \promptcomment|{question}@

  Task:
  Write the discussion for logic block. The current block must stay tightly focused on the question and on its own argument.

  Chunk pool for the current block:
  Chunk #i:
  \promptcomment|{evidence fragments}@

  Logic-block overview:
  Current logic block \promptcomment|{i}@:
  Argument: \promptcomment|{current argument}@
  Keywords: \promptcomment|{current keywords}@

  Logic block \promptcomment|{j}@:
  Argument: \promptcomment|{other argument}@
  Keywords: \promptcomment|{other keywords}@

  Output only the discussion for the current logic block, with inline citations.
  \imend
  \end{promptboxlist}

\setpromptboxtitle{Summary Stage Prompt Template (for deep-research style answers)}
\begin{promptboxlist}
  \promptcomment|//Subsection Summary Generation Prompt Template@
  \imstartsystem
  You are a professional answer-writing expert.
  You are writing \promptcomment|{one numbered subsection / one subsection}@ inside a larger report section.
  Your task is to produce the complete subsection, including the subsection title line and the subsection body, while preserving the factual support and inline citation markers already present in the source
  material.

  Requirements:
  1. Preserve valid inline citation markers such as [5] and [5,9] for claims that remain in the text.
  2. Do not invent new citation ids.
  3. Improve clarity, professionalism, and readability while keeping the content tightly focused on the question.
  4. If numbering is used, you must generate the subsection title yourself. If numbering is not used, do not generate an extra inner title.
  5. Output format is mandatory:
  \promptcomment|{subsection label}. Subsection Title
  Subsection body starts on the next line.@
  6. If earlier content already covers part of the same information, reduce repetition and keep only what is still necessary here.
  7. Every sentence must contribute directly to answering the question. Avoid open-ended expansion, tangents, and over-elaboration.
  8. \promptcomment|{first-line format requirement}@
  9. Output only the complete subsection with inline citations preserved.
  \imend

  \imstartuser
  Refine the current logic block into \promptcomment|{one numbered subsection / one subsection}@ inside its parent section.

  Question:
  \promptcomment|{question}@

  Current subsection label:
  \promptcomment|{subsection label}@.

  Current logic block information:
  Argument:
  \promptcomment|{argument}@

  Keywords:
  \promptcomment|{keywords}@

  Key information:
  \promptcomment|{key information}@

  Detailed content with inline citations:
  \promptcomment|{expanded discussion with citations}@

  Previously generated content:
  \promptcomment|{previous section headings and summaries}@

  Requirements:
  1. Preserve all valid citation markers already present in the detailed content, such as [5] or [5,9].
  2. Do not invent new citation ids, and do not delete citations that support retained claims.
  3. Improve clarity, density, and readability while keeping the content tightly focused on the question.
  4. If numbering is used, you must generate the subsection title yourself based on the current logic block. If numbering is not used, do not generate an extra inner title.
  5. Output format is mandatory and must be exactly:
  \promptcomment|{subsection label}. <your subsection title>
  <subsection content starts on the next line>@
  6. If the previously generated content already covers some information from the current logic block, reduce repeated discussion and keep only the incremental value needed here.
  7. Every part of the subsection must directly help answer the question. Stay problem-oriented and do not expand indefinitely.
  8. \promptcomment|{first-line format requirement}@
  9. Do not output any extra preface, explanation, or alternative formats.
  10. Make the subsection directly appendable after the previously generated content while maintaining coherence.
  11. Make the language clear, easy to understand, and highly readable, while avoiding excessive elaboration.
  12. The language you use should keep same with that of the question.

  Please generate the complete subsection for the current logic block:
  \imend
  \end{promptboxlist}

\setpromptboxtitle{Auxiliary Generation Prompt Templates (for deep-research style answers)}
\begin{promptboxlist}
  \promptcomment|//Direct Generation Prompt Template@
  \imstartsystem
  You are a professional report-writing expert. Your task is to discuss the argument of a specified logic block among a series of logic blocks, based on the provided literature passages and while staying
  closely focused on the question itself.

  Requirements:
  1. The discussion must stay closely aligned with the argument and the question, while also maintaining coherence with the already generated preceding answer.
  2. Refer to and cite information from the provided chunk pool comprehensively and extensively.
  3. Use concise and clear language.
  4. Maintain objectivity and accuracy.
  5. Staying close to the question is the first priority. Remove material that is only loosely related to the question.
  6. When a sentence uses information from the chunk pool, cite the local chunk ids at the end of the sentence in the format [1] or [1,2].
  7. Use only the chunk ids that appear in the prompt. Do not invent citation ids.

  Please generate a complete discussion and output the content directly with inline citations.
  \imend

  \imstartuser
  Please write a detailed discussion following instruction.

  Instruction:
  \promptcomment|{instruction}@

  Available evidence chunk pool for citation:
  Chunk #i:
  \promptcomment|{evidence fragments}@

  Requirements:
  1. Stay tightly focused on both the question and the argument.
  2. Use the chunk pool extensively when it is relevant.
  3. When a sentence uses information from a chunk, cite it with the provided local chunk id, such as [1] or [1,3].
  4. Use only the chunk ids shown in the chunk pool above. Do not invent citation ids.
  5. Place citations at the end of the corresponding sentence.
  6. Output only the discussion text.
  7. You must keep the language you use same with the question.

  Additional requirement for this direct-generation task:
  - Organize the answer using exactly two heading levels when structure is needed.
  - Use top-level headings in the form `A. ...`, `B. ...`, `C. ...`, `D. ...`.
  - Under each top-level heading, use second-level headings in the form `1. ...`, `2. ...`, `3. ...`, `4. ...` when there are multiple subpoints.
  - Keep the structure clear and reader-friendly.
  - Do not add any heading level deeper than the `1. / 2. / 3. / 4.` level.
  - Preserve inline citations in the generated content.
  - Keep the language of the discussion be the same with instruction (Chinese / English).
  \imend
  \promptsep
  \promptcomment|//Polish Prompt Template@
  \imstartsystem
  You are a careful editor for report sections.
  Polish the section for readability and concision while preserving all information and all valid inline citations.
  Return only the polished section text.
  \imend

  \imstartuser
  Instruction for this section:
  \promptcomment|{instruction}@

  Current section draft:
  \promptcomment|{section draft}@

  Editing requirements:
  1. Preserve all information that is already present in the draft.
  2. Preserve all valid inline citations such as [5] and [5,9]. Do not invent, remove, or renumber citations.
  3. Reduce useless discussion and repeated phrasing.
  4. Improve clarity, readability, and logical flow while preserving the original meaning, technical content, and conclusions.
  5. Remove redundancy and excessive elaboration. Merge overlapping ideas and keep the response focused on the target question.
  6. Strengthen the structure of the answer so that the argument follows a clear progression from background, to relationship/mechanism analysis, to modeling framework, and then to conclusions.
  7. Ensure that technical and domain-specific terms are accurate, consistent, and professionally used. Briefly clarify complex or ambiguous concepts when necessary.
  8. When discussing models, methods, or frameworks, clearly explain their purpose, assumptions, inputs, outputs, and how they connect to the main problem.
  9. Keep the writing tightly aligned with the instruction for this section.
  10. Keep the existing heading and numbering structure whenever it is already useful.
  11. Output only the polished section text. You don't need to add the section title.
  12. The language of your polished output should remain consistent with the draft.
  \imend
  \promptsep
  \promptcomment|//Answer Introduction Generation Prompt Template@
  \imstartuser
  Question:
  \promptcomment|{question}@

  Report body:
  \promptcomment|{assembled section contents}@

  Write the opening section of a report that answers the question.
  Requirements:
  1. Use 1-2 paragraphs.
  2. Introduce the question and briefly preview the report.
  3. Do not use bullet points.
  4. Do not include a heading.
  5. Do not include citation markers such as [1].
  6. Keep the language same with the report.
  \imend
  \promptsep
  \promptcomment|//Answer Conclusion Generation Prompt Template@
  \imstartuser
  Question:
  \promptcomment|{question}@

  Report body:
  \promptcomment|{assembled section contents}@

  Write the ending section of the report.
  Requirements:
  1. Use 1-2 paragraphs.
  2. Synthesize the key points and close the report.
  3. Do not introduce new section headings or bullet points.
  4. Do not include a heading.
  5. Follow the language of the question.
  6. Do not include citation markers such as [1].
  7. Keep the language same with the report.
  \imend
  \end{promptboxlist}

\twocolumn
\subsection{Case Study}
We present two examples to illustrate how DeepWeaver composes new claims by weaving related thought blocks across refinement steps.

\paragraph{LoQA} The question is how industrial water treatment should be redefined when shifting from ``pollution control'' to ``resource recovery.'' The initial claims discuss water reuse, energy recovery, material recovery, ecological evaluation, and economic feasibility separately. DeepWeaver links these claims with subordinate claims about circular resource use and multidimensional evaluation. The final result reframes wastewater treatment as a resource recovery system: wastewater is no longer treated only as pollution to remove, but as a resource pool for recovering water, energy, and valuable materials. This case shows DeepWeaver's ability to turn scattered technical points into a higher-level conceptual claim.

\paragraph{DeepResearch Bench} The topic is about the applications and functional capabilities of a sports intelligent tutoring system driven by multimodal data fusion. The draft claims cover real-time posture correction, performance evaluation, personalized training, injury prediction, and tactical analysis. DeepWeaver further connects these functions with signal-level, feature-level, and decision-level fusion. For example, posture correction is refined into a claim about fusing IMU and video data to build a 3D motion model for immediate feedback; performance evaluation is refined into feature-level fusion of biomechanical and physiological signals. This case shows how DeepWeaver converts separate application claims into a coherent set of system capabilities.

\section{Ethics Statement}

The source books used in this work are publicly published books and copyrighted works owned by their respective authors and publishers. They were accessed through our institution's paid and licensed electronic-book collections. We do not claim ownership over the original book text, and our access to these books does not imply any right to re-license or redistribute the original works.

For each book, we release only a limited excerpt of approximately 1,000 OCR-derived tokens, together with our derived annotations and metadata, for non-commercial academic research, reproducibility, and evaluation purposes. The release is neither intended nor sufficient to substitute for the original books, enable reconstruction of the original works, or circumvent the access restrictions of the licensed e-book collections.

We rely on fair-use-style research use for the limited excerpts, considering the non-commercial scholarly purpose, the limited amount of text released, and the absence of a market substitute for the original books. Users are responsible for ensuring that their use complies with applicable copyright laws and the terms of the original sources and institutional licenses. Upon request from copyright holders, we will review and remove disputed excerpts.

The data used in this work are derived from publicly published books and does not involve private user data, personal communications, or sensitive personal information. 

\section{Dataset License and Terms of Use}
\label{sec:license}

Annotations, metadata, and evaluation code in this work are released under Apache-2.0. Data are released under CC BY-NC 4.0, while the copyrighted book excerpts remain the property of their respective rights holders.

\onecolumn
\noindent

\noindent\textbf{Case Study on DeepResearch Bench}
 \begin{promptbox}{Inference Trajectory of DeepWeaver on DeepResearch Bench}
  Q:  Please write the section titled "III. Applications and Functional Capabilities" for a comprehensive report on the 'Construction and Application of a Sports Intelligent Tutoring and Learning Guidance System Driven by Multimodal Data Fusion.' This section must detail the practical applications and the specific functional capabilities of the proposed system. Explain how the system leverages multimodal data fusion to deliver intelligent tutoring and learning guidance in sports contexts, outlining its core features, use cases, and the value it provides to users such as athletes, coaches, or learners.

  \promptstep{Draft}
  \begin{promptverbatim}
  [
      {
        "argument": "The Sports Intelligent Tutoring and Learning Guidance System (SITLS) leverages multimodal data fusion to deliver real-time posture and movement correction by integrating diverse sensor inputs for precise biomechanical analysis.",
        "keywords": ["real-time correction", "multimodal fusion", "posture analysis", "inertial measurement units", "biomechanical feedback"],
        "key_info": "The SITLS uses a multi-sensor network including IMUs, pressure sensors, and vision-based cameras to capture high-resolution kinematic data. Through feature-level and decision-level fusion, the system combines these inputs to create a unified real-time representation of movement. This allows for the detection of deviations from optimal biomechanical patterns, such as knee valgus or inefficient swing planes, and provides immediate corrective feedback via audio, visual, or haptic signals.",
        "reference_chunks": [2, 4, 8, 9, 11, 14, 16, 21],
        "retrieval_reference_chunks": [97, 3, 23, 62, 63]
      },
      {
        "argument": "Comprehensive performance evaluation in the SITLS is enabled by fusing biomechanical, physiological, and visual data to assess athletic performance across multiple dimensions.",
        "keywords": ["performance evaluation", "multi-dimensional assessment", "biomechanical data", "physiological monitoring", "visual analysis"],
        "key_info": "The system integrates data from IMUs (capturing speed and acceleration), physiological sensors (monitoring heart rate variability and muscle oxygenation), and video analysis (for visual assessment of technique). This fusion allows for a holistic evaluation that goes beyond simple motion tracking to analyze the 'what', 'how', and 'why' of athletic performance, identifying subtle bottlenecks that single-modality systems would miss.",
        "reference_chunks": [2, 4, 10, 11, 14, 16, 19, 21],
        "retrieval_reference_chunks": [97, 104, 106, 109, 105]
      },
      {
        "argument": "Personalized training and learning pathways are dynamically generated by the SITLS through continuous updates to a student model based on individual performance data and learning preferences.",
        "keywords": ["personalized training", "adaptive learning", "student model", "machine learning", "dynamic adaptation"],
        "key_info": "The SITLS maintains a dynamic student model that updates in real-time based on performance data and feedback. By combining this model with a domain model of sports skills and a tutor model using machine learning algorithms, the system can identify individual strengths, weaknesses, and preferences to generate personalized training plans that adapt in real-time to ensure optimal challenge and relevance.",
        "reference_chunks": [2, 4, 8, 10, 14, 16, 19, 21],
        "retrieval_reference_chunks": [6, 3, 23, 20, 26]
      },
      {
        "argument": "Predictive injury risk assessment is achieved by analyzing long-term biomechanical patterns and real-time physiological signals to identify early warning signs of potential injury.",
        "keywords": ["injury prevention", "predictive analytics", "biomechanical load", "fatigue monitoring", "early warning system"],
        "key_info": "The SITLS integrates long-term biomechanical data (from IMUs and pressure sensors) with real-time physiological data (e.g., heart rate variability, muscle fatigue) and structural data (like CT scans) to detect deviations from an athlete's personal baseline. This enables the system to flag potential injury risks, such as asymmetry in joint loading or spikes in fatigue, long before a physical injury occurs, allowing for proactive intervention.",
        "reference_chunks": [10, 11, 14, 16, 19, 21, 23, 24],
        "retrieval_reference_chunks": [92, 61, 93, 75, 64]
      },
      {
        "argument": "In precision sports like golf and gymnastics, the SITLS provides unparalleled technical guidance by combining real-time sensor data with edge AI for immediate fault detection and correction.",
        "keywords": ["skill acquisition", "technical mastery", "golf training", "gymnastics analysis", "edge AI"],
        "key_info": "For golf, the system uses IMUs in gloves or clubs and cameras to analyze the entire swing sequence, detecting faults like early extension or reverse pivot in real-time. In gymnastics, it tracks rotation speed and body position during complex skills, providing feedback on air awareness and landing mechanics to ensure safety and technical accuracy. The use of edge AI enables immediate processing and feedback, allowing athletes to correct technique on the spot.",
        "reference_chunks": [8, 9, 14, 16, 19, 21, 24, 27],
        "retrieval_reference_chunks": [96, 97, 22, 95, 61]
      },
      {
        "argument": "In team sports, the SITLS enables collaborative dynamics analysis by fusing data from wearable sensors and positioning systems to quantify team coordination and tactical effectiveness.",
        "keywords": ["team coordination", "tactical analysis", "multi-agent systems", "positional data", "temporal synchronization"],
        "key_info": "The system integrates data from wearable sensors (IMUs, GPS) and positioning systems (LPS) to analyze collaborative dynamics across multiple players. It quantifies metrics like temporal coordination, spatial synchronization, and collective movement patterns, providing coaches with objective insights into team performance, such as identifying inconsistent defensive rotations or pressing intensity in soccer.",
        "reference_chunks": [4, 9, 14, 16, 19, 21, 24, 28],
        "retrieval_reference_chunks": [105, 104, 102, 106, 109]
      },
      ...,
      {
        "argument": "The SITLS provides coaches and learners with a powerful, data-driven decision-support tool that offers objective performance insights and personalized feedback.",
        "keywords": ["coaching support", "performance analytics", "data-driven decisions", "learner engagement", "objective feedback"],
        "key_info": "The system aggregates data from multiple athletes to provide coaches with a comprehensive view of team performance, workload, and injury risk, enabling data-informed decisions. For learners, it offers a personalized, engaging tutor with continuous, objective feedback that helps them understand progress, identify improvement areas, and build confidence through incremental successes.",
        "reference_chunks": [4, 9, 14, 16, 19, 21, 24, 28],
        "retrieval_reference_chunks": [20, 26, 23, 8, 75]
        }
    ]
    \end{promptverbatim}

  \promptstep{Subordinate (Refinement Turn \#1)}
  \begin{Verbatim}[
    breaklines=true,
    breakanywhere=true,
    breaksymbolleft={},
    breaksymbolright={},
    breakindent=0pt,
    breakautoindent=false,
    fontsize=\small
]
  [
      {
        "argument": "The Sports Intelligent Tutoring and Learning Guidance System (SITLS) integrates multimodal data to deliver real-time, precise posture and motion correction by fusing IMU kinematics with video-based pose estimation.",
        ...
      },
      {
        "argument": "Comprehensive performance evaluation in the SITLS is achieved through the integration of biomechanical, physiological, and movement efficiency metrics to provide a holistic assessment beyond traditional performance indicators.",
        ...
      },
      {
        "argument": "The SITLS generates personalized training and learning paths by dynamically adapting to an athlete's unique physiological and performance data, creating customized regimens that evolve in real time.",
        ...
      },
      {
        "argument": "Injury risk prediction and prevention in the SITLS is enabled by the fusion of biomechanical stress indicators with physiological fatigue signals, allowing for early detection of warning signs before injury occurs.",
        ...
      },
      {
        "argument": "The SITLS accelerates skill acquisition in complex sports by breaking down technical skills into constituent parts and providing targeted, real-time feedback, significantly reducing the learning curve.",
        ...
      },
      {
        "argument": "The SITLS enables advanced tactical analysis in team sports by fusing multi-player sensor data with video and game context, revealing patterns of coordination and cohesion that are invisible to traditional observation.",
        ...
      },
      ...,
      {
        "argument": "The SITLS enhances coach effectiveness by providing data-driven, objective assessments of athletes and offering personalized professional development recommendations based on their own teaching behaviors.",
        ...
      }
    ]
  \end{Verbatim}

  \promptstep{Commit (Refinement Turn \#1)}
  \begin{promptverbatim}
  {
      "trash": [
        {
          "argument": "The SITLS serves as a proactive health monitor by continuously tracking biomechanical load and fatigue, automatically adjusting training loads or recommending recovery protocols to prevent overuse injuries and optimize long-term athlete well-being.",
          ...
        },
        {
          "argument": "The SITLS enhances coach effectiveness by offering data-driven performance assessments and analyzing teaching behaviors through multimodal deep learning, providing personalized recommendations for professional development and improving pedagogical quality.",
          ...
        }
      ],
      "golden": [
        {
          "argument": "The Sports Intelligent Tutoring and Learning Guidance System (SITLS) delivers real-time, precise posture and movement correction by fusing IMU kinematics with video-based pose estimation to create a dynamic 3D model of athletic motion, enabling immediate detection and feedback on deviations in joint angles and movement trajectories.",
          ...
        },
        {
          "argument": "Comprehensive performance evaluation in the SITLS is achieved through the integration of biomechanical, physiological, and movement efficiency metrics, enabling a holistic assessment that captures not only performance outcomes but also underlying efficiency and fatigue indicators.",
          ...
        },
        {
          "argument": "Personalized training and learning pathways are dynamically generated by the SITLS through a real-time updated student model that adapts to individual physiological responses, performance data, and learning preferences using machine learning algorithms.",
          ...
        },
        {
          "argument": "Predictive injury risk assessment is enabled by fusing biomechanical stress indicators with physiological fatigue signals, allowing early detection of non-linear patterns such as asymmetrical joint loading and declining HRV before clinical injury occurs.",
          ...
        },
        {
          "argument": "In precision sports like golf and gymnastics, the SITLS provides real-time fault detection and technical correction using edge AI and multimodal sensor data, enabling immediate feedback on complex movement sequences such as swing mechanics and air awareness.",
          ...
        },
        {
          "argument": "In team sports, the SITLS enables advanced tactical analysis by fusing multi-player sensor data, positioning systems, and video with game context to generate spatiotemporal maps that quantify team coordination, synchronization, and collective movement patterns.",
          ...
        },
        ...,
        {
          "argument": "Advanced tactical analysis in team sports is enhanced by the SITLS through spatiotemporal mapping of multi-player dynamics, revealing hidden patterns of coordination, cohesion, and formation weaknesses that are invisible to traditional observation.",
          ...
        },
        {
          "argument": "The SITLS supports athlete rehabilitation by providing objective, quantifiable tracking of recovery progress through biomechanical and physiological metrics, ensuring safe return-to-play decisions and minimizing re-injury risk.",
          ...
        }
      ]
    }
  \end{promptverbatim}

  \promptstep{Subordinate (Refinement Turn \#2)}
  \begin{promptverbatim}
  [
      {
        "argument": "The SITLS achieves superior motion analysis by fusing raw sensor data at the signal level, reducing noise and motion artifacts to create a more accurate representation of athletic movement.",
        ...
      },
      {
        "argument": "Feature-level fusion enables the SITLS to extract and integrate high-dimensional, meaningful characteristics from diverse data streams into a unified multi-dimensional vector for comprehensive performance analysis.",
        ...
      },
      {
        "argument": "Decision-level fusion allows the SITLS to generate actionable, personalized recommendations by synthesizing fused data through advanced AI models like LSTM and Transformer networks.",
        ...
      },
      {
        "argument": "The SITLS provides real-time, multimodal feedback for posture and technique correction by combining wearable IMU data with video analysis to identify and correct movement deviations instantly.",
        ...
      },
      {
        "argument": "The SITLS enables personalized skill acquisition by integrating biomechanical and physiological data to evaluate performance and generate tailored training regimens based on individual weaknesses.",
        ...
      },
      {
        "argument": "The SITLS supports adaptive training optimization by continuously monitoring fatigue and physiological stress to adjust training intensity and recommend recovery protocols in real time.",
        ...
      },
      {
        "argument": "The SITLS enhances injury prevention by predicting risks through the correlation of long-term biomechanical and physiological data patterns, enabling proactive interventions.",
        ...
      },
      {
        "argument": "The SITLS facilitates advanced tactical analysis in team sports by fusing multi-player sensor data with video to visualize complex coordination patterns and team dynamics.",
        ...
      }
    ]
  \end{promptverbatim}

  \promptstep{Commit (Refinement Turn \#2)}
  \begin{promptverbatim}
  {
      "trash": [
        {
          "argument": "The SITLS serves as a proactive health monitor by continuously tracking biomechanical load and fatigue, automatically adjusting training loads or recommending recovery protocols to prevent overuse injuries and optimize long-term athlete well-being.",
          ...
        },
        {
          "argument": "The SITLS enhances coach effectiveness by offering data-driven performance assessments and analyzing teaching behaviors through multimodal deep learning, providing personalized recommendations for professional development and improving pedagogical quality.",
          ...
        },
        ...
      ],
      "golden": [
        {
          "argument": "The SITLS achieves highly accurate real-time motion analysis by fusing raw IMU and video data at the signal level, reducing noise and motion artifacts to create a robust, dynamic 3D model of athletic movement for precise deviation detection and immediate feedback.",
          ...
        },
        {
          "argument": "The system enables comprehensive performance evaluation by fusing biomechanical, physiological, and movement metrics at the feature level, extracting high-dimensional characteristics into a unified vector that captures the interplay between physical mechanics and biological state.",
          ...
        },
        {
          "argument": "Personalized training and learning pathways are dynamically generated through decision-level fusion of fused data using advanced AI models like LSTM and Transformer networks, enabling real-time, adaptive recommendations tailored to individual athlete profiles.",
          ...
        },
        {
          "argument": "Predictive injury risk assessment is enabled by analyzing long-term patterns in biomechanical stress and physiological fatigue, detecting early warning signs such as asymmetrical joint loading and declining HRV to support proactive intervention and injury prevention.",
          ...
        },
        {
          "argument": "Real-time technical correction in precision sports like golf and gymnastics is achieved by fusing IMU and video data to instantly detect faults in movement sequences—such as swing mechanics or air awareness—and deliver immediate multimodal feedback via audio, visual, or haptic cues.",
          ...
        },
        {
          "argument": "Advanced tactical analysis in team sports is enhanced by fusing multi-player sensor data with video to visualize spatiotemporal coordination patterns, enabling objective quantification of team dynamics, synchronization, and formation weaknesses.",
          ...
        },
        ...,
        {
          "argument": "The SITLS enables personalized skill acquisition by integrating biomechanical and physiological data to evaluate performance and generate tailored training regimens based on individual weaknesses.",
          ...
        },
        {
          "argument": "The SITLS supports adaptive training optimization by continuously monitoring fatigue and physiological stress to adjust training intensity and recommend recovery protocols in real time.",
          ...
        }
      ]
    }
  \end{promptverbatim}

  \promptstep{Expand}

  \begin{promptverbatim}
  [
      {
        "reference_blocks": [1, -1],
        "argument": "The SITLS achieves highly accurate real-time motion analysis by fusing raw IMU and video data at the signal level, reducing noise and motion artifacts to create a robust, dynamic 3D model of athletic movement for precise deviation detection and immediate feedback.",
        "keywords": ["real-time correction", "multimodal fusion", "posture analysis", "IMU", "video data", "3D motion modeling", "signal-level fusion", "motion analysis", "noise reduction", "sensor synchronization", "data robustness"],
        "key_info": "The SITLS uses a multi-sensor network including IMUs, pressure sensors, and vision-based cameras to capture high-resolution kinematic data. Through feature-level and decision-level fusion, the system combines these inputs to create a unified real-time representation of movement. ...",
        "reference_chunks": [2, 34, 4, 36, 70, 7, 8, 9, 11, 107, 14, 46, 16, 17, 78, 21, 91, 30],
        "retrieval_reference_chunks": [97, 96, 62, 106, 61],
        "response": "The Sports Intelligent Tutoring and Learning System (SITLS) achieves highly accurate real-time motion analysis by fusing raw Inertial Measurement Unit (IMU) and video data at the signal level, enabling a robust, dynamic 3D model of athletic movement that supports immediate feedback and precise deviation detection. Signal-level fusion combines unprocessed sensor streams—such as accelerometer, gyroscope, and magnetometer data from IMUs—with synchronized video frames at the most fundamental data level, allowing for enhanced noise reduction and motion artifact suppression before higher-level interpretation [78]. This approach overcomes limitations inherent in data-level fusion, where disparate sensor formats and noise profiles can lead to information conflict or redundancy, and decision-level fusion, which suffers from low information utilization and poor pattern mining [62]. By aligning raw IMU and video data through precise temporal synchronization, the SITLS constructs a temporally coherent 3D human skeleton model that reflects real-time kinematics with millimeter-level accuracy [91]. This integration enables the system to detect subtle deviations in posture, joint angles, and movement trajectories during high-speed or complex athletic actions, such as a golf swing or gymnastic dismount, where even minor errors can lead to performance degradation or injury risk. The resulting model serves as the foundational input for downstream functions, including real-time posture correction, performance evaluation, and technical feedback, ensuring that all intelligent tutoring outputs are grounded in a highly accurate and reliable representation of the athlete’s actual movement [62]. This capability is particularly critical in dynamic, open-ended sports domains where traditional single-modal systems fail to capture the full complexity of human motion, thereby enabling the SITLS to deliver data-driven, context-aware guidance that surpasses the limitations of conventional coaching and feedback mechanisms [2].",
        "citation": [78, 62, 91, 2]
      },
      ...
    ]
  \end{promptverbatim}

  \promptstep{Summary}
  \begin{promptverbatim}
{
    "summary": "A. Core Functional Capabilities\n1. Real-Time Posture Correction and Motion Analysis via Signal-Level Multimodal Fusion\nThe Sports Intelligent Tutoring and Learning System (SITLS) enables highly accurate real-time motion analysis by fusing raw Inertial Measurement Unit (IMU) and video data at the signal level, establishing a robust, dynamic 3D model of athletic movement for precise deviation detection and immediate feedback. This signal-level fusion integrates unprocessed IMU streams—including accelerometer, gyroscope, and magnetometer data—with synchronized video frames at the most fundamental data level, ensuring temporal coherence and minimizing noise and motion artifacts prior to higher-level interpretation [78]. By circumventing the limitations of data-level fusion—where mismatched sensor formats and noise profiles cause information conflict—and decision-level fusion—which suffers from reduced information utilization and weak pattern mining—the SITLS achieves superior data integration [62]. Precise temporal synchronization between IMU and video streams allows the system to construct a temporally consistent 3D human skeleton model with millimeter-level accuracy, capturing real-time kinematics during high-speed or complex movements such as a golf swing or gymnastic dismount [91]. This model facilitates the detection of subtle deviations in posture, joint angles, and movement trajectories, enabling early identification of biomechanical inefficiencies or injury risks. The resulting high-fidelity representation serves as the core input for downstream intelligent tutoring functions, including real-time posture correction, performance evaluation, and technical feedback, ensuring that guidance is both data-driven and context-aware. This capability is especially critical in dynamic, open-ended sports environments where single-modal systems fail to capture the full complexity of human motion, allowing the SITLS to deliver superior, adaptive coaching that surpasses traditional feedback mechanisms [2].\n\n2. Multi-Dimensional Performance Evaluation through Feature-Level Multimodal Fusion\nThe Sports Intelligent Tutoring and Learning System (SITLS) delivers a comprehensive, multi-dimensional performance evaluation by fusing biomechanical, physiological, and movement metrics at the feature level, transforming heterogeneous sensor data into a unified, high-dimensional vector that captures the dynamic interplay between physical mechanics and biological state. This capability is achieved through advanced feature-level fusion, which extracts modality-specific characteristics—such as joint angles, ground reaction forces, muscle activation patterns, heart rate variability (HRV), and lactate thresholds—from raw data streams before aligning them in a shared semantic space [62]. By processing features rather than raw signals or final decisions, the system avoids the pitfalls of data-level fusion—where mismatched noise profiles and sampling rates induce information conflict—and decision-level fusion, which underutilizes low-level patterns and degrades discriminative power [62,67]. This approach preserves rich, discriminative information while reducing redundancy and enhancing computational efficiency. The resulting multi-dimensional vector serves as a dynamic performance fingerprint, enabling real-time assessment of movement efficiency, energy expenditure, neuromuscular coordination, biomechanical symmetry, and fatigue onset. Integrated data from IMUs, wearable ECGs, force-sensitive insoles, and video-based motion capture allows the SITLS to detect subtle deviations in technique and early warning signs of overtraining or injury risk—capabilities unattainable by single-modality systems [91]. As demonstrated in related research, the fusion of structural and biomechanical data via deep learning significantly improves injury prediction accuracy, validating the value of integrated, multi-dimensional analysis [91]. In the SITLS, this fused representation not only informs immediate feedback but also supports longitudinal trajectory analysis, enabling coaches and athletes to optimize training loads, track skill progression, and refine long-term development strategies. The holistic assessment thus establishes a robust, data-driven foundation for intelligent tutoring, empowering users with deeper insights into execution quality, training readiness, and performance potential across all phases of athletic development.\n\n ...",
    "citation": [78, 62, 91, 2, 67, 1, 35, 96, 8, 102, 104, 105, 79, 75, 97, 73, 22, 26, 64, 4],
}
  \end{promptverbatim}

  \end{promptbox}

\begin{algorithm}[!htb]
\small
\caption{DeepWeaver Pseudocode}
\label{alg:deepweaver}
\begin{algorithmic}[1]
\Require Question $q$, TBC refinement rounds $n$, retrieval parameter $k$
\Require Number of evidence fragments per TBC generation $r$.
\Ensure Generated answer $y$.

\State Load question $q$, evidence pool $E=\{e_1,\ldots,e_m\}$ 
\State $\mathbf{H} \leftarrow \text{Emb}(e_1,e_2,\ldots,e_m)$
\State $E_r \leftarrow  \textsc{RandomSampling}(E,r)$
\State $\mathcal{T}_{M}^{(0)} \leftarrow \textsc{Draft}(q,E_r)$
\LineComment{Draft: Generate the initial main Thought Block Chain (TBC).}
\State $\mathcal{T}_{\mathrm{discard}} \leftarrow \emptyset$

\For{$t \leftarrow 0$ \textbf{to} $n-1$}
    \State $E_{\mathrm{cover}}
    \leftarrow \textsc{RetrieveCovered}(E,\mathbf{H},\mathcal{T}_{M}^{(t)},k)$

    \State $\mathcal{R}^{(t)}
    \leftarrow \textsc{RandomSampling}(E\setminus E_{\mathrm{cover}},r)$
    \LineComment{Collect evidence that is neglected or only weakly covered.}

    \State $\mathcal{T}_{S}^{(t)}
    \leftarrow \textsc{Subordinate}(q,\mathcal{R}^{(t)},\mathcal{T}_{\mathrm{discard}})$
    \LineComment{Subordinate: Construct a subordinate TBC from the residual evidence.}

    \State $\widetilde{\mathcal{T}}^{(t)}
    \leftarrow \textsc{Merge}(\mathcal{T}_{M}^{(t)},\mathcal{T}_{S}^{(t)})$
    \LineComment{Commit-Merge:  Commit useful subordinate claims into the main TBC.}

    \State $(\mathcal{T}_{\mathrm{discard}},\mathcal{T}_{M}^{(t+1)})
    \leftarrow \textsc{Discard}(q,\widetilde{\mathcal{T}}^{(t)},\mathcal{T}_{\mathrm{discard}})$
    \LineComment{Commit-Discard:  Remove redundant, irrelevant, or weakly supported blocks.}
\EndFor
\State $\mathcal{T}_{M}^{(n)}
\leftarrow \textsc{LinkEvidence}(T_M^{(n)}, \textsc{RetrieveCovered}(E,\mathbf{H},\mathcal{T}_{M}^{(n)},k))$
\LineComment{Link evidence with thought blocks based on top-k embedding retrieval.}
\State  $y_0\leftarrow \emptyset$
\ForAll{$b_i=(c_i,k_i,s_i,E_i) \in \mathcal{T}_{M}^{(n)}$}
    \State $S_i \leftarrow \textsc{Generate}(q,c_i,s_i,E_i)$
    \LineComment{Expand: Generate answer sections from each thought block.}
    \State $y_i \leftarrow  \textsc{Append}(y_{i-1},S_i)$
    \LineComment{Summary: Compose evidence-grounded sections into the final answer sequentially via LLM.}
\EndFor
\State  $y\leftarrow  y_{|\mathcal{T}_{M}^{(n)}|}$

\end{algorithmic}
\end{algorithm}

\end{document}